\pdfoutput=1

\documentclass[11pt]{article}

\usepackage[preprint]{acl}

\usepackage{times}
\usepackage{latexsym}

\usepackage[T1]{fontenc}

\usepackage[utf8]{inputenc}

\usepackage{microtype}

\usepackage{inconsolata}

\usepackage{graphicx}

\usepackage{stroop}
\usepackage[dvipsnames]{xcolor}
\usepackage{hyperref}
\usepackage{inconsolata}
\usepackage{tabularx}
\usepackage{listings,lstautogobble}
\usepackage{fancyvrb}
\usepackage{fvextra}
\usepackage{tikz}
\usepackage{amsmath}
\usetikzlibrary{positioning}
\usepackage{pgf}
\usepackage{comment}
\usepackage{arydshln}
\usepackage{color} 
\usepackage{colortbl}
\usepackage{longtable}
\usepackage{cuted}
\usepackage{fvextra}
\usepackage[most]{tcolorbox}
\tcbuselibrary{breakable}
\Crefname{tcolorbox}{Box}{Boxes}

\usepackage{siunitx}

\title{How Far Can Machine Translation Quality Take You? \\ Extrinsic Discourse Evaluation in Goal-Oriented Setups}

\author{Wafaa Mohammed  \qquad Kata Naszadi \qquad Vlad Niculae  \\
Language Technology Lab, University of Amsterdam \\
  \texttt{\{w.m.a.mohammed,k.naszadi,v.niculae\}@uva.nl}
}

\begin{document}
\maketitle

\begin{abstract}
Existing machine translation (MT) metrics and discourse-focused evaluations primarily assess translation quality intrinsically, without measuring the downstream consequences of translation errors. 
In this work, we focus on extrinsic discourse evaluation of machine translation under two distinct regimes: static and interactive. 
Under the static regime, we propose an entity counting task as a probe of referential consistency in discourse. 
We show that high intrinsic MT quality does not reliably predict downstream discourse success and strong MT systems still produce referential inconsistencies.
For the interactive regime, we study the goal-oriented multi-agent Welfare Diplomacy game as a probe of long-horizon communication and coordination. 
We find that interaction-specific translation failures impact downstream coordination.
Our results highlight goal-oriented environments as a viable framework for discourse-sensitive extrinsic MT evaluation.
\footnote{Code and data will be released upon acceptance.}

\end{abstract}

\section{Introduction}
Machine Translation evaluation has traditionally focused on intrinsically measuring the quality of generated translations in isolation \citep{freitag-etal-2024-llms}. 
However, MT systems are often deployed as intermediate components in larger pipelines such as business workflows or interpersonal communication, where their utility is determined by downstream task performance. 
Human-computer interaction research has studied the impact of machine translation when it mediates interpersonal communication \citep{DBLP:journals/pacmhci/ZhangOCG22},
highlighting in particular the importance of discourse \citep{DBLP:conf/cscw/YamashitaI06}.
Many discourse errors preserve local plausibility while altering cross-sentence interpretation, referential consistency, or interaction dynamics, making them difficult to capture with intrinsic metrics alone. However, discourse-focused MT evaluation has largely remained intrinsic, relying on targeted test suites and discourse-specific automatic metrics \citep{muller-etal-2018-large,jiang-etal-2023-discourse,zhao-etal-2023-discoscore,mohammed-etal-2024-findings}. Prior work has shown that standard MT metrics often correlate weakly with downstream performance \citep{moghe-etal-2023-extrinsic}, yet the downstream consequences of discourse-related mistranslations remain insufficiently understood.

In this work, we study extrinsic discourse evaluation of MT through goal-oriented environments, where translation quality is measured by its effect on downstream interpretation and coordination.
We examine two complementary settings: a static setting that isolates localized discourse interpretation failures, and an interactive regime that captures how failures propagate through sequential communication and coordination. Firstly, we design a controlled static setting of counting person entities in a text. 
This setting isolates referential consistency as the primary phenomenon of interest, which can be directly impacted by machine translation errors in grammatical correctness, anaphora resolution, and named entity translation, and allows us to directly measure how specific translation errors affect task success. 
Despite its simplicity, this setup captures core capabilities required in high-stakes applications such as legal interpretation, contractual liability, and threat assessment. 
Secondly, we study discourse in a dynamic multi-agent multi-turn communication setup. We utilize the strategic game Diplomacy, specifically the Welfare Diplomacy variant proposed by \citet{DBLP:journals/corr/abs-2310-08901}. This environment introduces long-horizon, interactive, and goal-driven communication, where translation errors can propagate, compound, and potentially influence outcomes.
Multi-agent communication games have grown increasingly sophisticated in recent years, with agent interactions covering non-trivial discourse phenomena such as complex anaphora resolution, conversational repair, and pragmatic implicature \citep{abdelnabi2023llm,horst2025playpen, zhou2024sotopia}.
These environments enable closed-loop evaluation for discourse in machine translation: the downstream impact of translation quality can be assessed automatically through game outcomes, and tracked across multiple conversational turns.

Using MT quality estimation metrics and qualitative evaluation, we uncover translation aspects that influence downstream outcomes. 
Our contributions can be summarized in the following:
\begin{itemize}
    \item We propose a framework for extrinsic discourse evaluation in MT based on two complementary goal-oriented regimes: a controlled static interpretation task and an interactive multi-agent communication environment.

    \item We introduce an entity counting setup that isolates discourse-level translation errors involving referential inconsistencies and coreference failure modes.

    \item We study multilingual communication in Diplomacy and show that intrinsic MT quality rankings transfer weakly to downstream coordination outcomes.

    \item Among the models evaluated, those with the highest COMETQE scores (eurollm 22B and ayaexpanse 8B) exhibit persistent discourse-level errors in coreference and coordination that impact downstream performance.

    \item We discuss the current state of MT adoption and outline directions and recommendations for moving toward more task-based evaluation frameworks.
\end{itemize}

\begin{table*}
\centering
\small
\begin{tabular}{c c}
\toprule
Text &  \# Entities  \\ 
\midrule
The architect thought that \colorbox{orange!20}{her} day was going well. The architect was happy to discuss \colorbox{orange!20}{her} career trajectory.  & 1 \\
The architect thought that \colorbox{orange!20}{her} day was going well. The architect was happy to discuss \colorbox{NavyBlue!20}{his} career trajectory.  & 2 \\
\bottomrule
\end{tabular}
\caption{Example entries from the entity counting task in \Cref{sec:entity_counting}.}
\label{entity_counting_example}
\end{table*}

\section{Entity Counting}
\label{sec:entity_counting}
The objective of this task is to determine the number of distinct individuals referenced in a given text. Although the task is intentionally controlled and structurally simple, it captures capabilities that are critical in high-stakes domains such as legal interpretation, contractual liability, and threat assessment. Real-world failures in this task are documented in news reports by \citet{bhuiyan2023lost,afghan}, which describe asylum applications being rejected when the single person pronoun ``I'' is mistranslated as ``we'', incorrectly implying that the application is for more than one person.
Unlike open-ended evaluations, the entity counting task provides a tightly controlled probe where downstream failure can be directly attributed to discourse-level translation distortions.

\subsection{Experimental Details}

\subsubsection{Data}
We build on the English dataset introduced by \citet{DBLP:journals/tacl/GautamBZLK24}, originally designed to evaluate pronoun fidelity, i.e., the ability to correctly reuse pronouns given an entity-pronoun pair introduced in context. All instances are unambiguous, with each pronoun having a single identifiable referent. To construct a controlled evaluation setting, we subsample the dataset according to the following criteria:

\begin{itemize}[topsep=0pt, partopsep=0pt, parsep=0pt, itemsep=0pt]
\item Each instance consists of a simple narrative involving a single individual, with a focus on pronominal reference.
\item We consider masculine (he/him/his) and feminine (she/her/her) third-person singular pronouns across nominative, accusative, and dependent possessive cases.\footnote{We focus on binary gendered pronouns to control for known model weaknesses related to gender stereotyping \citep{DBLP:journals/tacl/GautamBZLK24}. We acknowledge that this framing does not capture the full diversity of gender identities and pronoun usage, including singular ``they'' and neopronouns.}
\item No additional entities or distractors are introduced to the text.
\end{itemize}

The resulting subset comprises 360 instances. An example of the data is presented in \Cref{entity_counting_example}. Our preliminary experiments showed that large language models (LLMs) fail to produce consistent entity counts, exhibiting sensitivity to prompt phrasing and whether an explanation is requested. This is in line with reported unreliability of LLMs as judges \citep{wang-etal-2024-large-language-models-fair,chehbouni2026neither},
which undermines their usability for the task. Consequently, we rely on human annotation. Native speakers of the target languages, recruited as volunteers from our research group, are asked to annotate the dataset. Detailed annotation guidelines are in \Cref{app:annotation guidelines}.

\begin{table*}
\centering
\scalebox{0.9}{
\begin{tabular}{l c c c r r r}
\toprule
model & \multicolumn{3}{c}{COMETQE} & \multicolumn{3}{c}{Accuracy} \\
& Arabic & Russian & Romanian & Arabic & Russian & Romanian \\ 
\midrule
llama3.1 8B &  72.2 &  80.5 & 80.4 & 68.6 & \underline{84.7} & 79.4  \\
ayaexpanse 8B &  80.1 &  \underline{83.4} & 85.0 & 75.0 & 80.6  & 81.9 \\
gemma3 12B &  79.0 & 83.2 & \underline{85.4} & \underline{87.2} & \underline{88.6} & \underline{88.1} \\
eurollm 22B & \underline{81.0} &  \underline{83.7} & \underline{85.6} & 76.9 & 80.6  &  77.8 \\
\bottomrule
\end{tabular}
}
\caption{Translation performance on English$\rightarrow$Arabic, Russian, and Romanian along with human-evaluated entity counting accuracy. Statistically best values are \underline{underlined}. The table highlights that high quality translation does not correspond to high entity counting accuracy.}
\label{tab:entity_counting}
\end{table*}

\subsubsection{Models and languages\footnote{Huggingface repositories of all models used in this paper are in \Cref{app:model-repositories}}}
We experiment on Arabic, Russian and Romanian, covering diverse language families and writing systems to support generalizability. We evaluate the following translation models: \textbf{ayaexpanse 8B}: a LLM focusing on multilingual capabilities \citep{DBLP:journals/corr/abs-2412-04261}, \textbf{gemma3 12B}: a multi-modal vision-language model with wide language coverage and long context support \citep{DBLP:journals/corr/abs-2503-19786}, \textbf{llama3.1 8B}: a LLM optimized for multilingual dialogue use cases \citep{DBLP:journals/corr/abs-2407-21783}, and \textbf{eurollm 22B}: a LLM trained on EU languages \citep{DBLP:journals/corr/abs-2602-05879}.

\subsubsection{Evaluation}
In the absence of ground-truth references, we assess translation quality using the reference-free COMETQE metric \citep{chimoto-bassett-2022-comet}. For entity counting, we report accuracy based on correct entity counts.

\subsection{Results and Analysis}
\Cref{tab:entity_counting} reports translation and accuracy performance on entity counting across the three target languages. 
High COMETQE doesn’t guarantee correct gender agreement (eurollm 22B). Conversely, strong entity coherence can coexist with slightly lower COMETQE (gemma3 12B). These patterns suggest that while COMETQE likely captures translation dimensions regarding adequacy and fluency, entity counting accuracy reflects discourse-level coherence and coreference consistency, where models may score well on COMETQE yet still make referent-splitting mistakes.

\begin{figure}[t]
\centering
\includegraphics[width=\columnwidth]{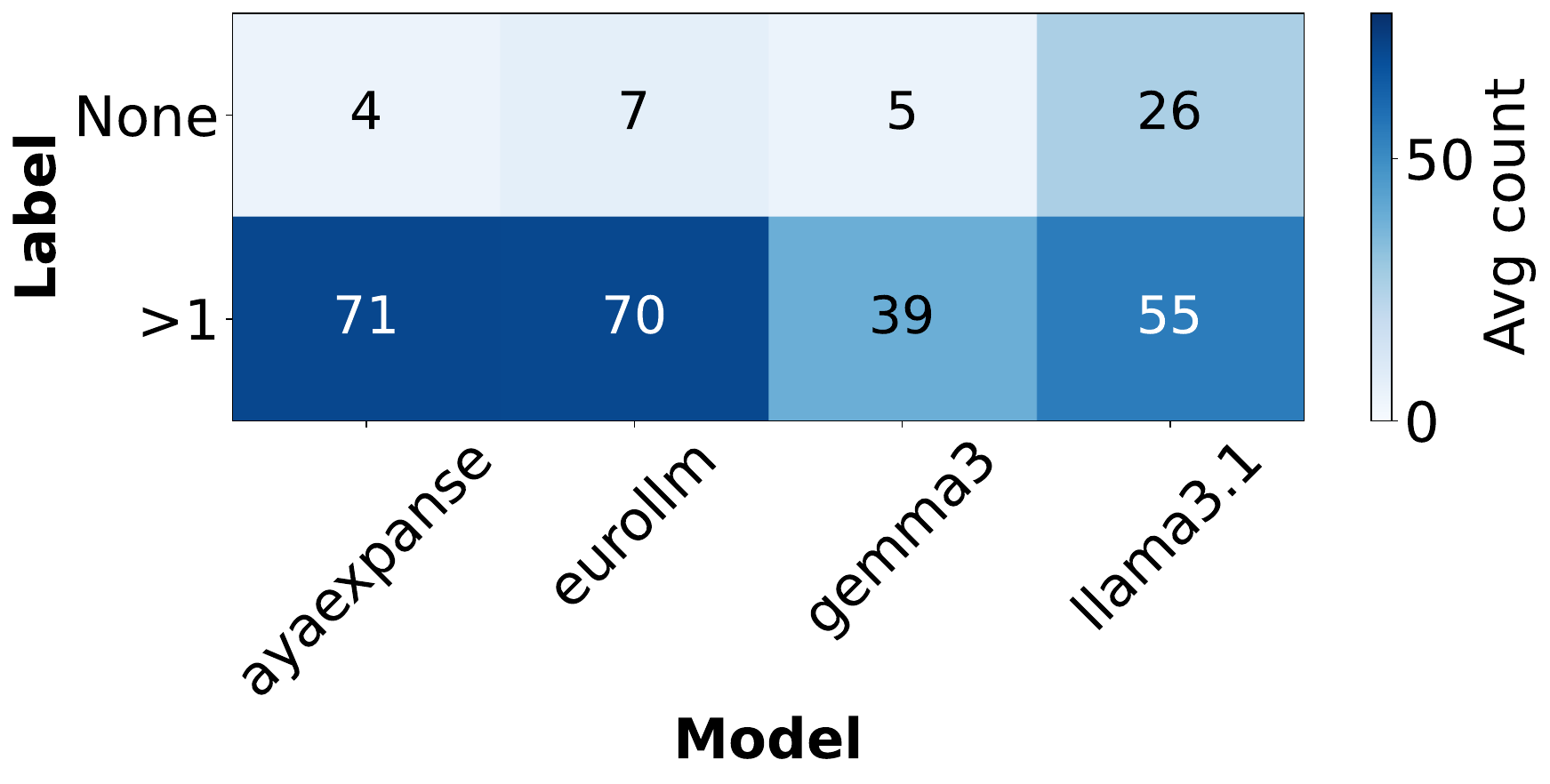}
\caption{Entities proportions across models averaged across target languages. The plot shows that top performing translation models (eurollm and ayaexpanse) still frequently make coreference resolution errors.}
\label{fig:proportions}
\end{figure}

\Cref{fig:proportions} shows the distribution of erroneous entity counts across model outputs.The ground truth reference is \textit{one} entity per sample. Through a manual qualitative analysis, we find that deviations in entity counts >1 indicate two error patterns. The first is a gender flip in pronoun resolution (i.e., using a masculine pronoun for a feminine referent or vice versa), which can cause the pronoun to be interpreted as referring to a different entity. The second is inconsistent lexical realization of the entity, where the same referent is expressed with alternating masculine and feminine noun forms across the output. 
The \textit{None} label indicates ungrammatical/indeterminate cases (mismatched gender morphology between verb and noun alongside inconsistent pronoun agreement) where entity count can’t be reliably determined. Most models’ main failure mode is >1 entities, which remains unexpectedly frequent even in high-performing COMETQE models (ayaexpanse and eurollm). llama3.1 is uniquely impacted by \textit{None} cases, suggesting more frequent agreement or grammatical breakdowns which is associated with lower perceived translation quality.  

Overall, entity counting provides a complementary signal, exposing task-relevant failures not captured by standard translation quality metrics. The task-specific analysis highlights distinct failure modes and model-specific error patterns.

\begin{figure*}
    \centering
    \begin{subfigure}{0.48\textwidth}
        \centering
        \includegraphics[width=\textwidth]{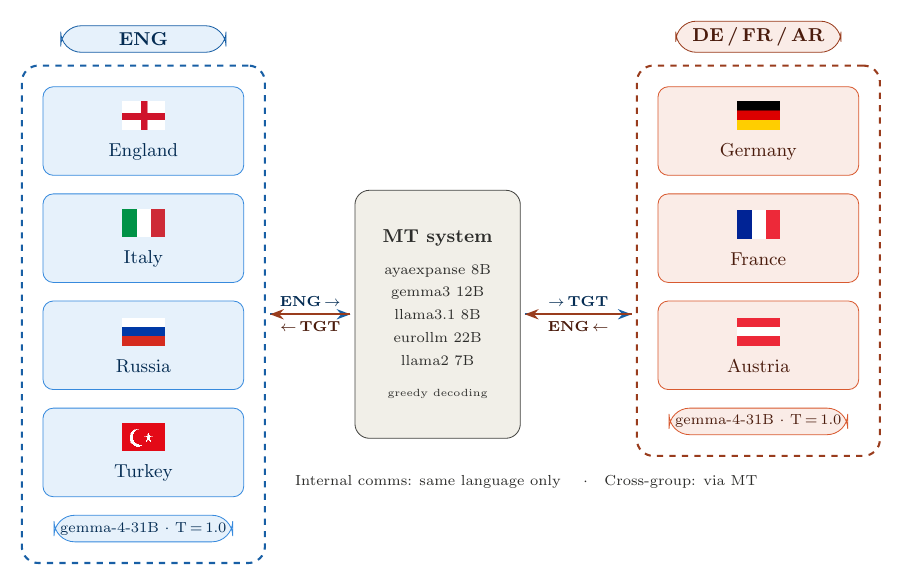}
    \end{subfigure}
    \hfill
    \begin{subfigure}{0.48\textwidth}
        \centering
        \includegraphics[width=.85\textwidth]{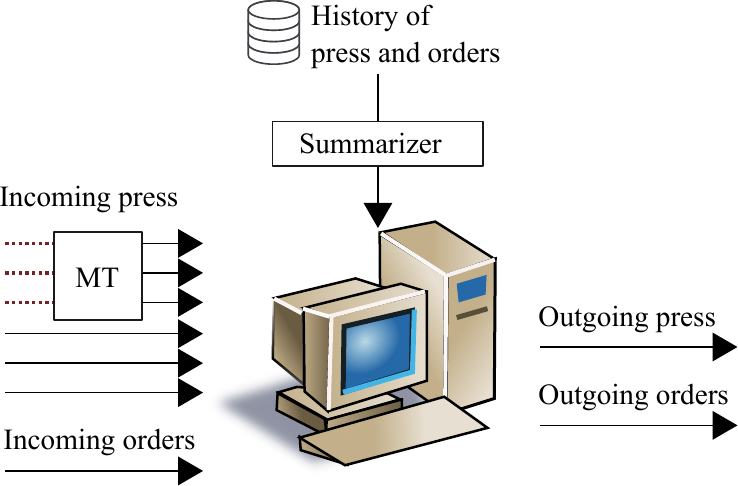}
    \end{subfigure}

    \caption{Left: an overview of the multilingual communication setup between all players. Right: a detailed view from the perspective of one player. Dotted lines are messages in a foreign language; all other solid lines are in the language of the player considered.}
    \label{fig:multilingual-game}
\end{figure*}

\section{Multi-Agent Communication}
We study the game \textit{Diplomacy}, specifically the \textbf{Welfare Diplomacy (WD)} variant proposed by \citet{DBLP:journals/corr/abs-2310-08901}, in which players must balance military expansion with domestic welfare. In standard Diplomacy (SD), seven players control military units with the objective of capturing supply centers (SCs). The game proceeds in yearly cycles consisting of Spring, Fall, and Winter phases.
During Spring and Fall, players issue orders to their units. In Winter, players may build new units if their SC count has increased over the preceding year, or must disband units if it has decreased. 
The game ends when a player controls 18 SCs. In the ``full-press'' SD setup, players can exchange private messages prior to issuing orders in each phase. WD introduces three key modifications to this setup: (1) after each Winter phase, players receive Welfare Points (WPs) equal to the difference between the number of SCs they control and their number of units on the board; (2) the game terminates after a fixed number of years instead of a victory threshold; and (3) each player aims to maximize their own WPs, meaning there is no single winner. We experiment on WD under the following constraints: the game lasts four years, with three phases per year, and five communication rounds per phase. During communication rounds, players exchange messages but do not explicitly reveal order-level details. Orders are submitted only at the end of each phase and are executed immediately.

\begin{figure*}
\centering
\includegraphics[width=\textwidth]{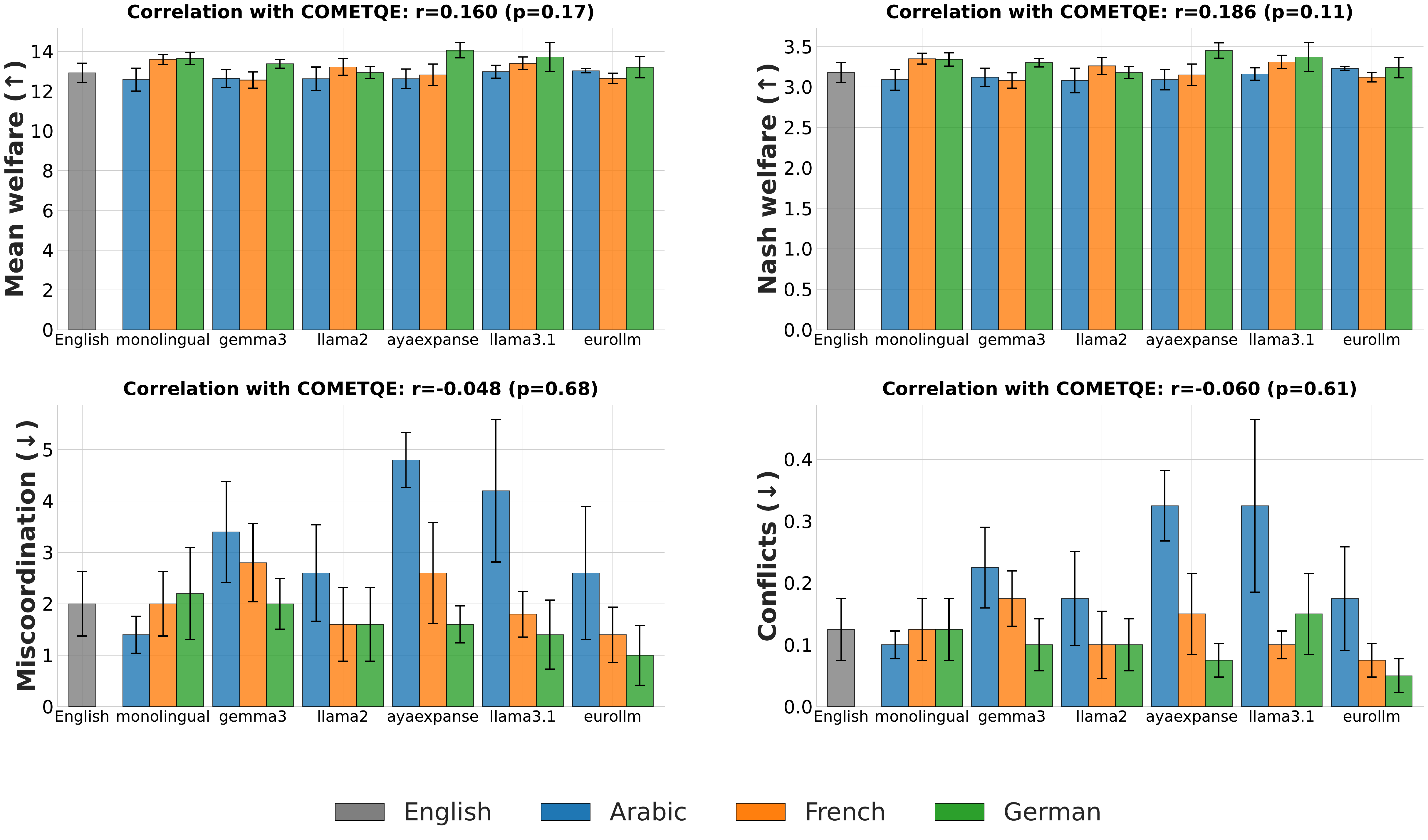}
\caption{Game performance metrics: Nash and mean welfare, miscoordination, and average conflicts in the monolingual (all communication is in target language), multilingual using different translation models, and English baseline setups. The bars show mean scores and standard errors across 5 runs. The plot also highlights the Pearson correlation of game metrics to COMETQE scores in multilingual setups. Correlations are computed over 75 points (3 target languages $\times$ 5 translation models $\times$ 5 game runs). Order of columns matches the order of the legend items.} 
\label{fig:task_metrics}
\end{figure*}

\subsection{Experimental Details}

\subsubsection{Communication setups}
We consider four communication settings:
\begin{itemize}
    \item \textbf{Monolingual communication:} All players communicate in a single language (English or one of the target languages). This serves as a baseline to verify that LLM agents are competent in each language, ensuring that observed errors in the multilingual setting stem from translation rather than difference of game performance based on language.
    
    \item \textbf{Multilingual communication:} Players are partitioned into two groups, each using a different language. Each player only reads and writes messages in its assigned language. Communication between groups is mediated by a translation system, which translates messages across languages (see \Cref{fig:multilingual-game}).

    \item \textbf{No-press baseline:} No messages are exchanged throughout the game and players only issue orders. This enables studying communication's effect on the game's outcome.

    \item \textbf{Random communication baseline:} Players exchange random messages (noise). This enables comparing meaningful communication to non-informative interaction. We use the same random policy setup as that of \citet{DBLP:journals/corr/abs-2310-08901}.
\end{itemize}
To facilitate reproducability, we provide examples of different game components in \Cref{app:game_prompt_details}, including system and user prompts, summarization and translation prompts, as well as of the random message baseline.

\subsubsection{Models and languages}
The game includes seven powers: Germany, France, Austria, England, Italy, Russia, and Turkey, all simulated by a gemma-4-31B agent. In the multilingual setup, powers are split into two groups: \{England, Italy, Russia, Turkey\} communicating in English and \{Germany, France, Austria\} communicating in the target language. We keep the assignment of powers to language groups fixed across all experiments to eliminate the variance introduced by different power permutations. We leave it to future work to explore different permutations.
We experiment on three target languages: German, French, and Arabic.
For translation between the groups, we test five models: ayaexpanse 8B, gemma3 12B, llama3.1 8B, eurollm 22B, and llama2 7B \citep{DBLP:journals/corr/abs-2307-09288}. Player agents use temperature sampling (T=1.0, top-p=0.95, top-k=64), while translation models use greedy decoding. Each configuration is run five times with different seeds.

\subsubsection{Evaluation}
Similar to the entity counting task, we evaluate translation quality using reference-free COMETQE metric. For game performance, we report four measures: \textbf{mean welfare}: average welfare points across all powers at the end of the game, capturing overall prosperity, \textbf{Nash welfare}: geometric mean of yearly welfare per power, reflecting fairness and cooperative balance, \textbf{miscoordination}: number of invalid or failed orders including bounced, void, and no convoy actions, and \textbf{average conflicts}: mean number of contested tiles where multiple units attempt to move or hold simultaneously.


\subsection{Results and Analysis}

\begin{figure}
\centering
\includegraphics[width=\columnwidth]{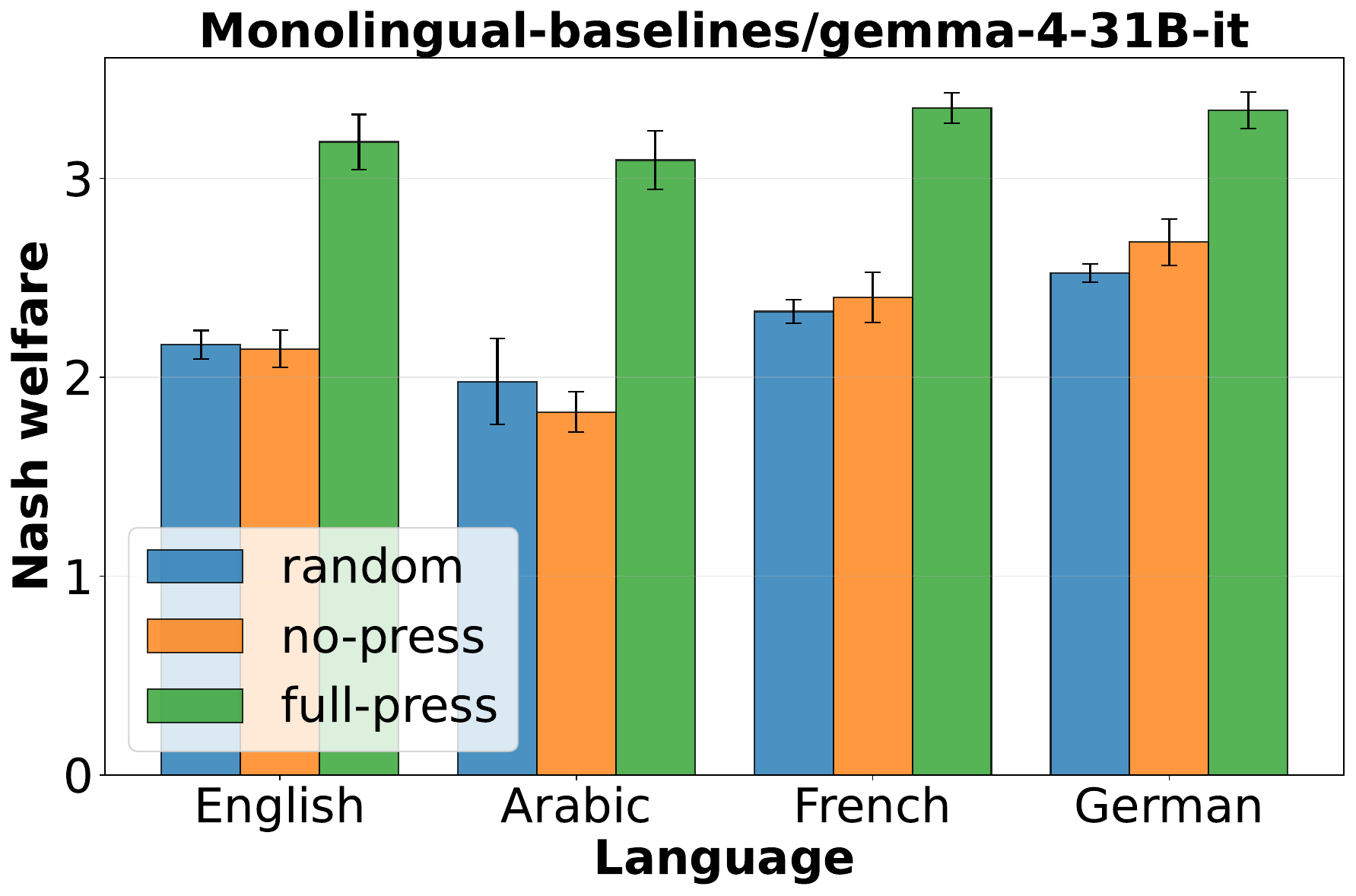}
\caption{
Nash welfare value across monolingual, no-press, and random communication setups. The plot highlights the value of meaningful communication, and the agent's competence across languages. Left-to-right columns match top-to-bottom order of legend items.
}
\label{fig:baselines}
\end{figure}

\paragraph{Communication is essential to global social welfare.} \Cref{fig:baselines} compares full-press monolingual communication with no-press and random press setups. Results demonstrate that communication is important, and specifically meaningful communication. It also confirms that the agent model is competent in all target languages.

\paragraph{Translation affects how welfare is distributed.} 
\Cref{tab:cometqe_scores} reports translation performance across target languages. \Cref{fig:task_metrics} shows game performance under monolingual and multilingual setups using different translation models, along with Pearson correlations between game and translation metrics. The English monolingual setup provides a strong baseline, with high welfare and low miscoordination and conflict rates. Welfare metrics (mean and Nash welfare) remain relatively stable across languages and setups, whereas coordination metrics (miscoordination and conflicts) show larger variation, indicating that translation primarily affects interaction dynamics rather than aggregate utility. In terms of translation quality in \Cref{tab:cometqe_scores}, eurollm 22B, gemma3 12B, and ayaexpanse 8B achieve comparable COMETQE scores, while llama2 7B and llama3.1 8B underperform. However, this ranking does not transfer to welfare and coordination outcomes. This mismatch is reflected in weak Pearson correlations between COMETQE and game metrics (\Cref{fig:task_metrics}), 
particularly for miscoordination and conflicts, suggesting that standard MT quality metrics are poor predictors of coordination behavior. Effects vary by language: in Arabic and French, translation tends to degrade coordination relative to the monolingual setting, whereas German shows the opposite pattern, with translation improving coordination outcomes. Further analysis (next paragraph) suggests that this effect is associated with increased formality introduced by German-to-English translation. 

\begin{table}
\centering
\begin{tabular}{l c c c}
\toprule
Model & Ar & Fr & De \\
\midrule
llama2 7B & 55.2 & 74.4 & 68.8 \\
llama3.1 8B &  74.8 & 78.8 & 78.8 \\
ayaexpanse 8B & 78.9 & 79.2 & \textbf{80.3} \\
gemma3 12B & 79.5 & 79.3 & 79.1 \\
eurollm 22B & \textbf{80.2} & \textbf{81.5} & 79.8  \\
\bottomrule
\end{tabular}
\caption{COMETQE on English$\rightarrow$Arabic(Ar), French(Fr), and German(De), averaged over 5 runs for each model. \textbf{Bold} highlights best values per column.}
\label{tab:cometqe_scores}
\end{table}

\begin{table*}
\sisetup{table-format=1.2}
\centering
\begin{tabular}{l r S S S S S}
\toprule
phenomenon & total & {COMETQE} & {miscoord.} & {conflicts} & {welfare mean} & {Nash} \\
\midrule
length & 31 & -0.83$^\ddagger$ & 0.07 & 0.08 & -0.01 & -0.03 \\
preface & 2142 & -0.93$^\ddagger$ & 0.03 & 0.04 & -0.11 & -0.13 \\
digit & 79 & -0.89$^\ddagger$ & 0.06 & 0.07 & -0.09 & -0.11 \\
mood & 29 & -0.77$^\ddagger$ & -0.10 & -0.11 & -0.23 & -0.26 \\
acronyms & 2216 & -0.45$^\ddagger$ & 0.32$^\dagger$ & 0.28 & -0.26 & -0.28 \\
dis/arm & 46 & 0.08 & 0.32$^\dagger$ & 0.34$^\dagger$ & -0.04 & -0.03  \\
formality & 323 & 0.09 & -0.27 & -0.25 & 0.27 & 0.27 \\
\bottomrule
\end{tabular}
\caption{Pearson correlation coefficients between discourse phenomena and translation and game metrics. Correlations are computed over 75 points (3 target languages $\times$ 5 translation models $\times$ 5 game runs). Statistically significant correlations are marked with $^\dagger$p< 0.05, $^\ddagger$p< 0.01 (FDR-adjusted for multiple comparisons). The table highlight that COMETQE is more sensitive to surface level phenomena (first four), while game metrics correlate more with game-specific phenomena (last three).}
\label{tab:error_analysis}
\end{table*}

\paragraph{Discourse phenomena and their correlation with overall translation and game performance.} While a comprehensive manual analysis of all message-level phenomena is infeasible at scale, we use several automatically identifiable proxies to study error patterns in the messages exchanged during the game. Examples of each phenomenon are in \Cref{app:error_examples}:
\begin{itemize}
    \item \textbf{Length mismatch:} Discrepancy in informational density between source and target (length ratio in words $>$ 3.0 or $<$ 0.25).
    \item \textbf{Assistant preface:} Presence of meta-communicative or procedural framing in the target (“sure, here is the translation”, “here is the translation”).
    \item \textbf{Digit mismatch:} Inconsistencies in 4-digit numerical strings (e.g., 1901) between source and target, reflecting failures in preserving referential anchors.
    \item \textbf{Mood change:} Source and target differ in the sentence mood (measured by the number of question marks "?"), which impacts negotiation dynamics and information elicitation.
    \item \textbf{Acronym mismatch:} Inconsistent rendering of uppercase 2-5 letter acronyms (e.g., map tokens like ENG, NTH, or game acronyms like WP) between source and target, disrupting shared reference and introducing ambiguity in strategic communication.
    \item \textbf{Disarmament mistranslation:} Cases where \textit{disarmament} is mistranslated as its antonym \textit{armament}, affecting strategic interpretation.
    \item \textbf{Formality increase:} Shifts from informal (e.g., ``Let's stick to the deal'') to formal (e.g., ``It is in our mutual interest'') register between original and translated English messages, measured via formality lexicons (\Cref{app:formality_lexicon}). This phenomena uniquely affects English-German multilingual setups.
\end{itemize}

We expect these discourse phenomena to vary in their impact on downstream outcomes.
In \Cref{tab:error_analysis}, we report the Pearson correlations between the identified errors and both translation and game metrics. The correlations are computed over 75 points (3 languages $\times$ 5 translation models $\times$ 5 game runs). We observe that COMETQE is highly sensitive to some surface- and token-level distortions (e.g., length mismatch, assistant preface, digit mismatch, and mood change), but comparatively less sensitive to game-specific discourse phenomena such as acronym mismatch, disarmament mistranslation and formality.
In contrast, these latter phenomena exhibit stronger correlations with game performance metrics, suggesting a closer association with interactive success.
Overall, our results highlight important discourse aspects that are not adequately captured by standard MT quality metrics, despite having a measurable impact on downstream task performance.

\section{Discussion}

\paragraph{MT-mediated interaction has become mainstream.} Machine translation increasingly mediates everyday communication between speakers of different languages. It is routinely embedded in extended, goal-oriented interactions such as customer support, collaborative work, travel coordination, and social communication \citep{DBLP:conf/cscw/YamashitaI06,DBLP:journals/pacmhci/ZhangOCG22,mohammed-etal-2024-findings}, where conversations unfold across multiple turns. In these settings, translation errors can accumulate over time, distort shared understanding, and affect the interaction outcome. This raises the importance of discourse-level phenomena, including coherence, consistency, referential accuracy, and pragmatic alignment, which traditional MT evaluation metrics are not designed to capture. Therefore, evaluating MT in interactive, multilingual settings requires moving beyond intrinsic metrics toward frameworks that capture how translation quality influences interaction dynamics and task outcomes.

\paragraph{Strong MT systems fail at basic referential consistency in discourse.} Even in a maximally controlled setting, strong MT systems struggle with maintaining consistent reference to a single entity across a short text. This is particularly striking, as it establishes a surprisingly weak baseline for discourse-level reliability. While modern MT systems achieve strong performance on evaluation benchmarks, these results suggest that basic forms of contextual consistency are not adequately captured by existing overall quality metrics. This gap is especially concerning in interactive settings, where referential inconsistencies can compound over multiple turns and interfere with shared understanding between participants.

\paragraph{LLM communication games for goal-oriented MT evaluation.} LLM-based agents have demonstrated strong performance in multi-agent communication games, suggesting that modern language models are capable participants and reliable communicators in complex interactive settings \citep{xia2024measuring}. These LLM-to-LLM communication games offer a new path toward automated and goal-grounded evaluation by combining the efficiency of automated evaluation with task-anchored interaction signals. It allows translation quality to be assessed in terms of its effect on goal achievement rather than isolated linguistic proxies.

\paragraph{Shifts between intrinsic and extrinsic MT evaluation.} Early MT research prioritized external, application-based evaluation in which translation quality is judged through performance on downstream tasks carried out by human users in realistic settings (see \Cref{related-work}). Over time, evaluation shifted toward intrinsic, metric-based approaches, a trend reinforced by the rise of large language models and scalable benchmarking paradigms such as LLM-as-a-judge, which made automatic evaluation more practical and reproducible \citep{kocmi-etal-2025-findings}. As discussed by \citet{reiter-2025-evaluate}, this shift is also driven by structural factors in the field: limited expertise in impact-oriented evaluation, the benchmark-centric machine learning culture, and the practical difficulty and cost of conducting real deployment studies. Recent work has begun to challenge this paradigm, arguing that effective evaluation should be anchored in the needs of those deploying and using translation systems and integrated into development pipelines that explicitly account for external constraints and objectives, rather than being detached from application contexts \citep{kayano-sugawara-2025-specification,saphra-etal-2024-first,DBLP:journals/corr/abs-2602-24055}.

\section{Related Work}
\label{related-work}
 
\paragraph{Discourse evaluation in MT.} Studying discourse characteristics in MT has been going on since the early works on rule-based and statistical machine translation systems \citep{hardmeier2012discourse}. Works involve creating discourse-specific datasets \citep{jiang-etal-2023-discourse,DBLP:journals/corr/abs-2004-14607,DBLP:journals/corr/abs-2210-14667}, developing model architectures to specifically target discourse translation \citep{bawden-etal-2018-evaluating,tiedemann-scherrer-2017-neural}, designing evaluation frameworks and metrics targeting discourse performance \citep{DBLP:journals/corr/abs-2208-09118,bawden-etal-2018-evaluating,zhao-etal-2023-discoscore}, and interpreting the role of different parts of the input in discourse translation \citep{mohammed-niculae-2024-measuring,mohammed2024context}.

\paragraph{Multi-agent communication.} 
An increasing number of works propose using community of Large Language Models (LLMs) rather than a single agent to solve complex tasks. For a comprehensive, communication-focused review of LLM-based multi-agent systems (LLM-MAS), see \cite{yan2025beyond}. Most influential to our work are systems that utilize natural language as a core tool for inter-agent communication. When evaluating how machine translation impacts goal achievement, tasks requiring free-form, multi-turn communication are especially compelling, as they naturally introduce discourse-related challenges. For instance, \cite{liang2024encouraging} introduce multi-agent debate to improve arithmetic reasoning, where the mixed incentives of the participants spark rich natural language exchanges across successive turns. Similarly, \cite{chen2025optima} deliberately introduce information asymmetry in multi-hop question-answering, which necessitates strategic discourse planning as agents must intentionally coordinate what information to solicit from one another across turns. Furthermore, multi-agent communities containing more than two agents with distinct roles or personas  also encourage highly dynamic and intricate language use \citep{wang2024unleashing, hong2024metagpt, park2023generative}.

\paragraph{Extrinsic machine translation evaluation.} Task-based MT evaluation has evolved from early extrinsic frameworks linking translation quality to downstream task performance such as extraction, filtering, and multilingual processing \citep{white1998task,article}. Subsequent human-based evaluations measure MT usefulness through information extraction tasks, quizzes, proficiency tests, and comprehension assessments \citep{voss-tate-2006-task,DBLP:journals/pbml/BerkaCB11,matsuzaki-etal-2015-evaluating,jones-etal-2007-ilr}. More recent work introduces toolkits and automated benchmarks for reading comprehension, cross-lingual information retrieval, and LLM-based QA evaluation \citep{scarton-specia-2016-reading,sun-etal-2020-clireval, fernandes2025llms,zhang-etal-2025-litransproqa}, alongside application-specific studies in e-commerce, medical systems, toxicity detection, triage, and political classification \citep{zhang-etal-2022-evaluating,zocca-etal-2025-experience,bell-etal-2025-translate,martindale-carpuat-2025-improving,osorio-etal-2025-devil}.

\section{Conclusion}
In this work, we investigate the usefulness of machine translation in goal oriented applications. We study discourse characteristics of translation in both static and interactive regimes and examine their impact on the downstream task performance. Our results indicate a weak correlation between translation metrics (COMETQE) and downstream performance in both regimes. We further identify several task-specific discourse phenomena such as coreference resolution, named entity translation, negation, and formality that are not adequately captured by COMETQE, and that influence downstream coreference and coordination performance. 
Future research could extend this discourse-centric analysis to a broader range of downstream applications of machine translation. Another promising direction is the development of domain-- and task-specific automated evaluation metrics that better capture discourse effectiveness and better correlate with downstream utility.

\section*{Limitations}
\begin{itemize}
    \item In the discourse analysis in the Welfare Diplomacy game, we investigate the correlation of the different phenomena to the translation and game performance. We emphasize that our findings are purely correlational and should not be interpreted as evidence of causal relationships between specific phenomena and observed performance outcomes.
    \item To isolate the effect of the translation quality, we do not present any explicit task-specific context or instructions to the translation models. Future work could study how to improve translations with task specific instructions.
    \item The entity counting task is subject to annotator bias and implicit assumptions during labeling. In particular, linguistic and cultural conventions may influence annotation decisions; for example, in some languages it is traditionally acceptable to use masculine forms when referring to occupational nouns with feminine referents \citep{alenazy2024arabic}. 
    \item We rely on a single metric for machine translation evaluation, namely COMETQE. Although COMETQE is a well established reference-free metric for assessing translation quality, other evaluation metrics (such as metricX \citep{juraska-etal-2023-metricx,juraska-etal-2024-metricx}) may capture complementary aspects of translation quality and could yield different relative patterns across systems or conditions.
\end{itemize}

\section*{Ethical Considerations}

As discussed in the paper, machine translation is increasingly deployed in high-stakes settings such as legal communication, asylum procedures, and healthcare access, where translation errors may affect users’ ability to make informed decisions or complete critical tasks. Our evaluation framework focuses on discourse-sensitive phenomena relevant to goal-oriented interaction, but does not capture many important aspects of real-world communication, including cultural nuance, sociolinguistic variation, demographic bias, or domain-specific institutional requirements. We believe that meaningful evaluation of deployed MT systems requires collaboration with external stakeholders, including professional translators, domain experts, institutions, and affected communities.

\section*{Acknowledgements}
We would like to thank Evgeniia Tokarchuk for her assistance with the annotations. We would also like to thank Sergey Troshin and Vagrant Gautam for their valuable discussions and feedback on this work.
This research was funded in part by the Netherlands Organization for Scientific Research (NWO) under project numbers VI.Veni.212.228. KN received funding from the Hybrid Intelligence Center, a 10-year programme funded by the Dutch Ministry of Education, Culture and Science through the
Netherlands Organisation for Scientific Research
with grant number 024.004.022.
This work used the Dutch national supercomputer Snellius with the support of the SURF Cooperative and NWO funding using grant no.\ 2023.017.

\bibliography{anthology-1,anthology-2,custom}

\begin{thebibliography}{61}
\providecommand{\natexlab}[1]{#1}

\bibitem[{Abdelnabi et~al.(2023)Abdelnabi, Gomaa, Sivaprasad, Sch{\"o}nherr, and Fritz}]{abdelnabi2023llm}
Sahar Abdelnabi, Amr Gomaa, Sarath Sivaprasad, Lea Sch{\"o}nherr, and Mario Fritz. 2023.
\newblock Llm-deliberation: Evaluating llms with interactive multi-agent negotiation game.

\bibitem[{Alenazy et~al.(2024)Alenazy, Almahameed, Tawalbeh, and Abu-Elrob}]{alenazy2024arabic}
Mamdouh~A Alenazy, Yazan~S Almahameed, Ahmad~I Tawalbeh, and Rula Abu-Elrob. 2024.
\newblock Arabic profession-denoting nouns and gender: A morphosyntactic analysis of agreement.
\newblock \emph{Theory and Practice in Language Studies}, 14(12):3733--3743.

\bibitem[{Bawden et~al.(2018)Bawden, Sennrich, Birch, and Haddow}]{bawden-etal-2018-evaluating}
Rachel Bawden, Rico Sennrich, Alexandra Birch, and Barry Haddow. 2018.
\newblock \href {https://doi.org/10.18653/v1/N18-1118} {Evaluating discourse phenomena in neural machine translation}.
\newblock In \emph{Proceedings of the 2018 Conference of the North {A}merican Chapter of the Association for Computational Linguistics: Human Language Technologies, Volume 1 (Long Papers)}, pages 1304--1313, New Orleans, Louisiana. Association for Computational Linguistics.

\bibitem[{Bell et~al.(2025)Bell, S{\'a}nchez, Dale, Stenetorp, Artetxe, and Costa-Juss{\`a}}]{bell-etal-2025-translate}
Samuel Bell, Eduardo S{\'a}nchez, David Dale, Pontus Stenetorp, Mikel Artetxe, and Marta~R. Costa-Juss{\`a}. 2025.
\newblock \href {https://doi.org/10.18653/v1/2025.wmt-1.15} {Translate, then detect: Leveraging machine translation for cross-lingual toxicity classification}.
\newblock In \emph{Proceedings of the Tenth Conference on Machine Translation}, pages 253--268, Suzhou, China. Association for Computational Linguistics.

\bibitem[{Berka et~al.(2011)Berka, Cern{\'{y}}, and Bojar}]{DBLP:journals/pbml/BerkaCB11}
Jan Berka, Martin Cern{\'{y}}, and Ondrej Bojar. 2011.
\newblock \href {http://ufal.mff.cuni.cz/pbml/95/art-berka-cerny-bojar.pdf} {Quiz-based evaluation of machine translation}.
\newblock \emph{Prague Bull. Math. Linguistics}, 95:77--86.

\bibitem[{Bhuiyan(2023)}]{bhuiyan2023lost}
Johana Bhuiyan. 2023.
\newblock \href {https://www.theguardian.com/us-news/2023/sep/07/asylum-seekers-ai-translation-apps} {Lost in ai translation: Growing reliance on language apps jeopardizes some asylum applications}.
\newblock \emph{The Guardian}, 7.

\bibitem[{Chehbouni et~al.(2026)Chehbouni, Haddou, Cheung, and Farnadi}]{chehbouni2026neither}
Khaoula Chehbouni, Mohammed Haddou, Jackie~CK Cheung, and Golnoosh Farnadi. 2026.
\newblock Neither valid nor reliable? investigating the use of llms as judges.
\newblock \emph{Advances in Neural Information Processing Systems}, 38.

\bibitem[{Chen et~al.(2025)Chen, Yuan, Qian, Yang, Liu, and Sun}]{chen2025optima}
Weize Chen, Jiarui Yuan, Chen Qian, Cheng Yang, Zhiyuan Liu, and Maosong Sun. 2025.
\newblock \href {https://doi.org/10.18653/v1/2025.findings-acl.601} {Optima: Optimizing effectiveness and efficiency for {LLM}-based multi-agent system}.
\newblock In \emph{Findings of the Association for Computational Linguistics: ACL 2025}, pages 11534--11557, Vienna, Austria. Association for Computational Linguistics.

\bibitem[{Chimoto and Bassett(2022)}]{chimoto-bassett-2022-comet}
Everlyn~Asiko Chimoto and Bruce~A. Bassett. 2022.
\newblock \href {https://doi.org/10.18653/v1/2022.findings-emnlp.348} {{COMET}-{QE} and active learning for low-resource machine translation}.
\newblock In \emph{Findings of the Association for Computational Linguistics: EMNLP 2022}, pages 4735--4740, Abu Dhabi, United Arab Emirates. Association for Computational Linguistics.

\bibitem[{Dang et~al.(2024)Dang, Singh, D'souza, Ahmadian, Salamanca, Smith, Peppin, Hong, Govindassamy, Zhao, Kublik, Amer, Aryabumi, Campos, Tan, Kocmi, Strub, Grinsztajn, Flet{-}Berliac, Locatelli, Lin, Talupuru, Venkitesh, Cairuz, Yang, Chung, Ko, Shi, Shukayev, Bae, Piktus, Castagn{\'{e}}, Cruz{-}Salinas, Kim, Crawhall{-}Stein, Morisot, Roy, Blunsom, Zhang, Gomez, Frosst, Fadaee, Ermis, {\"{U}}st{\"{u}}n, and Hooker}]{DBLP:journals/corr/abs-2412-04261}
John Dang, Shivalika Singh, Daniel D'souza, Arash Ahmadian, Alejandro Salamanca, Madeline Smith, Aidan Peppin, Sungjin Hong, Manoj Govindassamy, Terrence Zhao, Sandra Kublik, Meor Amer, Viraat Aryabumi, Jon~Ander Campos, Yi~Chern Tan, Tom Kocmi, Florian Strub, Nathan Grinsztajn, Yannis Flet{-}Berliac, and 26 others. 2024.
\newblock \href {https://doi.org/10.48550/ARXIV.2412.04261} {Aya expanse: Combining research breakthroughs for a new multilingual frontier}.
\newblock \emph{CoRR}, abs/2412.04261.

\bibitem[{Deck(2023)}]{afghan}
Andrew Deck. 2023.
\newblock \href {https://restofworld.org/2023/ai-translation-errors-afghan-refugees-asylum/} {Ai translation is jeopardizing afghan asylum claims}.
\newblock \emph{Rest of World}.

\bibitem[{Doyon et~al.(1999)Doyon, White, and Taylor}]{article}
Jennifer Doyon, John White, and Kathryn Taylor. 1999.
\newblock Task-based evaluation for machine translation.

\bibitem[{Fernandes et~al.(2025)Fernandes, Agrawal, Zaranis, Martins, and Neubig}]{fernandes2025llms}
Patrick Fernandes, Sweta Agrawal, Emmanouil Zaranis, Andr{\'e}~FT Martins, and Graham Neubig. 2025.
\newblock Do llms understand your translations? evaluating paragraph-level mt with question answering.
\newblock \emph{arXiv preprint arXiv:2504.07583}.

\bibitem[{Freitag et~al.(2024)Freitag, Mathur, Deutsch, Lo, Avramidis, Rei, Thompson, Blain, Kocmi, Wang, Adelani, Buchicchio, Zerva, and Lavie}]{freitag-etal-2024-llms}
Markus Freitag, Nitika Mathur, Daniel Deutsch, Chi-Kiu Lo, Eleftherios Avramidis, Ricardo Rei, Brian Thompson, Frederic Blain, Tom Kocmi, Jiayi Wang, David~Ifeoluwa Adelani, Marianna Buchicchio, Chrysoula Zerva, and Alon Lavie. 2024.
\newblock \href {https://doi.org/10.18653/v1/2024.wmt-1.2} {Are {LLM}s breaking {MT} metrics? results of the {WMT}24 metrics shared task}.
\newblock In \emph{Proceedings of the Ninth Conference on Machine Translation}, pages 47--81, Miami, Florida, USA. Association for Computational Linguistics.

\bibitem[{Gautam et~al.(2024)Gautam, Bingert, Zhu, Lauscher, and Klakow}]{DBLP:journals/tacl/GautamBZLK24}
Vagrant Gautam, Eileen Bingert, Dawei Zhu, Anne Lauscher, and Dietrich Klakow. 2024.
\newblock \href {https://doi.org/10.1162/TACL\_A\_00719} {Robust pronoun fidelity with english llms: Are they reasoning, repeating, or just biased?}
\newblock \emph{Trans. Assoc. Comput. Linguistics}, 12:1755--1779.

\bibitem[{Hardmeier(2012)}]{hardmeier2012discourse}
Christian Hardmeier. 2012.
\newblock Discourse in statistical machine translation. a survey and a case study.
\newblock \emph{Discours. Revue de linguistique, psycholinguistique et informatique. A journal of linguistics, psycholinguistics and computational linguistics}, (11).

\bibitem[{Hong et~al.(2024)Hong, Zhuge, Chen, Zheng, Cheng, Wang, Zhang, Yau, Lin, Zhou et~al.}]{hong2024metagpt}
Sirui Hong, Mingchen Zhuge, Jonathan Chen, Xiawu Zheng, Yuheng Cheng, Jinlin Wang, Ceyao Zhang, Steven Yau, Zijuan Lin, Liyang Zhou, and 1 others. 2024.
\newblock Metagpt: Meta programming for a multi-agent collaborative framework.
\newblock In \emph{International Conference on Learning Representations}, volume 2024, pages 23247--23275.

\bibitem[{Horst et~al.(2025)Horst, Mazzaccara, Schmidt, Sullivan, Moment{\`e}, Franceschetti, Sadler, Hakimov, Testoni, Bernardi et~al.}]{horst2025playpen}
Nicola Horst, Davide Mazzaccara, Antonia Schmidt, Michael Sullivan, Filippo Moment{\`e}, Luca Franceschetti, Philipp Sadler, Sherzod Hakimov, Alberto Testoni, Raffaella Bernardi, and 1 others. 2025.
\newblock Playpen: An environment for exploring learning from dialogue game feedback.
\newblock In \emph{Proceedings of the 2025 Conference on Empirical Methods in Natural Language Processing}, pages 29842--29879.

\bibitem[{Jiang et~al.(2022)Jiang, Liu, Ma, Zhang, Sachan, and Cotterell}]{DBLP:journals/corr/abs-2210-14667}
Yuchen~Eleanor Jiang, Tianyu Liu, Shuming Ma, Dongdong Zhang, Mrinmaya Sachan, and Ryan Cotterell. 2022.
\newblock \href {https://doi.org/10.48550/ARXIV.2210.14667} {A bilingual parallel corpus with discourse annotations}.
\newblock \emph{CoRR}, abs/2210.14667.

\bibitem[{Jiang et~al.(2023)Jiang, Liu, Ma, Zhang, Sachan, and Cotterell}]{jiang-etal-2023-discourse}
Yuchen~Eleanor Jiang, Tianyu Liu, Shuming Ma, Dongdong Zhang, Mrinmaya Sachan, and Ryan Cotterell. 2023.
\newblock \href {https://doi.org/10.18653/v1/2023.acl-long.435} {Discourse-centric evaluation of document-level machine translation with a new densely annotated parallel corpus of novels}.
\newblock In \emph{Proceedings of the 61st Annual Meeting of the Association for Computational Linguistics (Volume 1: Long Papers)}, pages 7853--7872, Toronto, Canada. Association for Computational Linguistics.

\bibitem[{Jones et~al.(2007)Jones, Herzog, Ibrahim, Jairam, Shen, Gibson, and Emonts}]{jones-etal-2007-ilr}
Douglas Jones, Martha Herzog, Hussny Ibrahim, Arvind Jairam, Wade Shen, Edward Gibson, and Michael Emonts. 2007.
\newblock \href {https://aclanthology.org/N07-2020/} {{ILR}-based {MT} comprehension test with multi-level questions}.
\newblock In \emph{Human Language Technologies 2007: The Conference of the North {A}merican Chapter of the Association for Computational Linguistics; Companion Volume, Short Papers}, pages 77--80, Rochester, New York. Association for Computational Linguistics.

\bibitem[{Juraska et~al.(2024)Juraska, Deutsch, Finkelstein, and Freitag}]{juraska-etal-2024-metricx}
Juraj Juraska, Daniel Deutsch, Mara Finkelstein, and Markus Freitag. 2024.
\newblock \href {https://doi.org/10.18653/v1/2024.wmt-1.35} {{M}etric{X}-24: The {G}oogle submission to the {WMT} 2024 metrics shared task}.
\newblock In \emph{Proceedings of the Ninth Conference on Machine Translation}, pages 492--504, Miami, Florida, USA. Association for Computational Linguistics.

\bibitem[{Juraska et~al.(2023)Juraska, Finkelstein, Deutsch, Siddhant, Mirzazadeh, and Freitag}]{juraska-etal-2023-metricx}
Juraj Juraska, Mara Finkelstein, Daniel Deutsch, Aditya Siddhant, Mehdi Mirzazadeh, and Markus Freitag. 2023.
\newblock \href {https://doi.org/10.18653/v1/2023.wmt-1.63} {{M}etric{X}-23: The {G}oogle submission to the {WMT} 2023 metrics shared task}.
\newblock In \emph{Proceedings of the Eighth Conference on Machine Translation}, pages 756--767, Singapore. Association for Computational Linguistics.

\bibitem[{Jwalapuram et~al.(2020)Jwalapuram, Rychalska, Joty, and Basaj}]{DBLP:journals/corr/abs-2004-14607}
Prathyusha Jwalapuram, Barbara Rychalska, Shafiq~R. Joty, and Dominika Basaj. 2020.
\newblock \href {https://arxiv.org/abs/2004.14607} {Can your context-aware {MT} system pass the dip benchmark tests? : Evaluation benchmarks for discourse phenomena in machine translation}.
\newblock \emph{CoRR}, abs/2004.14607.

\bibitem[{Kayano and Sugawara(2025)}]{kayano-sugawara-2025-specification}
Yoko Kayano and Saku Sugawara. 2025.
\newblock \href {https://doi.org/10.18653/v1/2025.wmt-1.7} {Specification-aware machine translation and evaluation for purpose alignment}.
\newblock In \emph{Proceedings of the Tenth Conference on Machine Translation}, pages 113--141, Suzhou, China. Association for Computational Linguistics.

\bibitem[{Kocmi et~al.(2025)Kocmi, Artemova, Avramidis, Bawden, Bojar, Dranch, Dvorkovich, Dukanov, Fishel, Freitag, Gowda, Grundkiewicz, Haddow, Karpinska, Koehn, Lakougna, Lundin, Monz, Murray, Nagata, Perrella, Proietti, Popel, Popovi{\'c}, Riley, Shmatova, Steingr{\'i}msson, Yankovskaya, and Zouhar}]{kocmi-etal-2025-findings}
Tom Kocmi, Ekaterina Artemova, Eleftherios Avramidis, Rachel Bawden, Ond{\v{r}}ej Bojar, Konstantin Dranch, Anton Dvorkovich, Sergey Dukanov, Mark Fishel, Markus Freitag, Thamme Gowda, Roman Grundkiewicz, Barry Haddow, Marzena Karpinska, Philipp Koehn, Howard Lakougna, Jessica Lundin, Christof Monz, Kenton Murray, and 10 others. 2025.
\newblock \href {https://doi.org/10.18653/v1/2025.wmt-1.22} {Findings of the {WMT}25 general machine translation shared task: Time to stop evaluating on easy test sets}.
\newblock In \emph{Proceedings of the Tenth Conference on Machine Translation}, pages 355--413, Suzhou, China. Association for Computational Linguistics.

\bibitem[{Liang et~al.(2024)Liang, He, Jiao, Wang, Wang, Wang, Yang, Shi, and Tu}]{liang2024encouraging}
Tian Liang, Zhiwei He, Wenxiang Jiao, Xing Wang, Yan Wang, Rui Wang, Yujiu Yang, Shuming Shi, and Zhaopeng Tu. 2024.
\newblock \href {https://doi.org/10.18653/v1/2024.emnlp-main.992} {Encouraging divergent thinking in large language models through multi-agent debate}.
\newblock In \emph{Proceedings of the 2024 Conference on Empirical Methods in Natural Language Processing}, pages 17889--17904, Miami, Florida, USA. Association for Computational Linguistics.

\bibitem[{Martindale and Carpuat(2025)}]{martindale-carpuat-2025-improving}
Marianna~J. Martindale and Marine Carpuat. 2025.
\newblock \href {https://aclanthology.org/2025.mtsummit-1.45/} {Improving {MT}-enabled triage performance with multiple {MT} outputs}.
\newblock In \emph{Proceedings of Machine Translation Summit XX: Volume 1}, pages 592--607, Geneva, Switzerland. European Association for Machine Translation.

\bibitem[{Matsuzaki et~al.(2015)Matsuzaki, Fujita, Todo, and Arai}]{matsuzaki-etal-2015-evaluating}
Takuya Matsuzaki, Akira Fujita, Naoya Todo, and Noriko~H. Arai. 2015.
\newblock \href {https://doi.org/10.3115/v1/P15-2024} {Evaluating machine translation systems with second language proficiency tests}.
\newblock In \emph{Proceedings of the 53rd Annual Meeting of the Association for Computational Linguistics and the 7th International Joint Conference on Natural Language Processing (Volume 2: Short Papers)}, pages 145--149, Beijing, China. Association for Computational Linguistics.

\bibitem[{Moghe et~al.(2023)Moghe, Sherborne, Steedman, and Birch}]{moghe-etal-2023-extrinsic}
Nikita Moghe, Tom Sherborne, Mark Steedman, and Alexandra Birch. 2023.
\newblock \href {https://doi.org/10.18653/v1/2023.acl-long.730} {Extrinsic evaluation of machine translation metrics}.
\newblock In \emph{Proceedings of the 61st Annual Meeting of the Association for Computational Linguistics (Volume 1: Long Papers)}, pages 13060--13078, Toronto, Canada. Association for Computational Linguistics.

\bibitem[{Mohammed et~al.(2024)Mohammed, Agrawal, Farajian, Cabarr{\~a}o, Eikema, Farinha, and C.~De~Souza}]{mohammed-etal-2024-findings}
Wafaa Mohammed, Sweta Agrawal, Amin Farajian, Vera Cabarr{\~a}o, Bryan Eikema, Ana~C Farinha, and Jos{\'e}~G. C.~De~Souza. 2024.
\newblock \href {https://doi.org/10.18653/v1/2024.wmt-1.59} {Findings of the {WMT} 2024 shared task on chat translation}.
\newblock In \emph{Proceedings of the Ninth Conference on Machine Translation}, pages 701--714, Miami, Florida, USA. Association for Computational Linguistics.

\bibitem[{Mohammed and Niculae(2024{\natexlab{a}})}]{mohammed2024context}
Wafaa Mohammed and Vlad Niculae. 2024{\natexlab{a}}.
\newblock Context-aware or context-insensitive? assessing llms' performance in document-level translation.
\newblock \emph{arXiv preprint arXiv:2410.14391}.

\bibitem[{Mohammed and Niculae(2024{\natexlab{b}})}]{mohammed-niculae-2024-measuring}
Wafaa Mohammed and Vlad Niculae. 2024{\natexlab{b}}.
\newblock \href {https://aclanthology.org/2024.findings-eacl.113/} {On measuring context utilization in document-level {MT} systems}.
\newblock In \emph{Findings of the Association for Computational Linguistics: EACL 2024}, pages 1633--1643, St. Julian{'}s, Malta. Association for Computational Linguistics.

\bibitem[{Mukobi et~al.(2023)Mukobi, Erlebach, Lauffer, Hammond, Chan, and Clifton}]{DBLP:journals/corr/abs-2310-08901}
Gabriel Mukobi, Hannah Erlebach, Niklas Lauffer, Lewis Hammond, Alan Chan, and Jesse Clifton. 2023.
\newblock \href {https://doi.org/10.48550/ARXIV.2310.08901} {Welfare diplomacy: Benchmarking language model cooperation}.
\newblock \emph{CoRR}, abs/2310.08901.

\bibitem[{M{\"u}ller et~al.(2018)M{\"u}ller, Rios, Voita, and Sennrich}]{muller-etal-2018-large}
Mathias M{\"u}ller, Annette Rios, Elena Voita, and Rico Sennrich. 2018.
\newblock \href {https://doi.org/10.18653/v1/W18-6307} {A large-scale test set for the evaluation of context-aware pronoun translation in neural machine translation}.
\newblock In \emph{Proceedings of the Third Conference on Machine Translation: Research Papers}, pages 61--72, Brussels, Belgium. Association for Computational Linguistics.

\bibitem[{Osorio et~al.(2025)Osorio, Alshammari, Alatrush, Heintze, Converse, Alsarra, Khan, Brandt, and D{'}Orazio}]{osorio-etal-2025-devil}
Javier Osorio, Afraa Alshammari, Naif Alatrush, Dagmar Heintze, Amber Converse, Sultan Alsarra, Latifur Khan, Patrick~T. Brandt, and Vito D{'}Orazio. 2025.
\newblock \href {https://aclanthology.org/2025.mtsummit-1.24/} {The devil is in the details: Assessing the effects of machine-translation on {LLM} performance in domain-specific texts}.
\newblock In \emph{Proceedings of Machine Translation Summit XX: Volume 1}, pages 315--332, Geneva, Switzerland. European Association for Machine Translation.

\bibitem[{Park et~al.(2023)Park, O'Brien, Cai, Morris, Liang, and Bernstein}]{park2023generative}
Joon~Sung Park, Joseph O'Brien, Carrie~Jun Cai, Meredith~Ringel Morris, Percy Liang, and Michael~S. Bernstein. 2023.
\newblock \href {https://doi.org/10.1145/3586183.3606763} {Generative agents: Interactive simulacra of human behavior}.
\newblock In \emph{Proceedings of the 36th Annual ACM Symposium on User Interface Software and Technology}, UIST '23, New York, NY, USA. Association for Computing Machinery.

\bibitem[{Ramos et~al.(2026)Ramos, Alves, Gisserot{-}Boukhlef, Alves, Martins, Fernandes, Pombal, Guerreiro, Rei, Boizard, Farajian, Klimaszewski, de~Souza, Haddow, Yvon, Colombo, Birch, and Martins}]{DBLP:journals/corr/abs-2602-05879}
Miguel~Moura Ramos, Duarte~M. Alves, Hippolyte Gisserot{-}Boukhlef, Jo{\~{a}}o Alves, Pedro~Henrique Martins, Patrick Fernandes, Jos{\'{e}} Pombal, Nuno~Miguel Guerreiro, Ricardo Rei, Nicolas Boizard, M.~Amin Farajian, Mateusz Klimaszewski, Jos{\'{e}} G.~C. de~Souza, Barry Haddow, Fran{\c{c}}ois Yvon, Pierre Colombo, Alexandra Birch, and Andr{\'{e}} F.~T. Martins. 2026.
\newblock \href {https://doi.org/10.48550/ARXIV.2602.05879} {Eurollm-22b: Technical report}.
\newblock \emph{CoRR}, abs/2602.05879.

\bibitem[{Reiter(2025)}]{reiter-2025-evaluate}
Ehud Reiter. 2025.
\newblock \href {https://doi.org/10.1162/coli.a.18} {We should evaluate real-world impact}.
\newblock \emph{Computational Linguistics}, 51(4):1419--1431.

\bibitem[{Saphra et~al.(2024)Saphra, Fleisig, Cho, and Lopez}]{saphra-etal-2024-first}
Naomi Saphra, Eve Fleisig, Kyunghyun Cho, and Adam Lopez. 2024.
\newblock \href {https://doi.org/10.18653/v1/2024.naacl-long.128} {First tragedy, then parse: History repeats itself in the new era of large language models}.
\newblock In \emph{Proceedings of the 2024 Conference of the North American Chapter of the Association for Computational Linguistics: Human Language Technologies (Volume 1: Long Papers)}, pages 2310--2326, Mexico City, Mexico. Association for Computational Linguistics.

\bibitem[{Scarton and Specia(2016)}]{scarton-specia-2016-reading}
Carolina Scarton and Lucia Specia. 2016.
\newblock \href {https://aclanthology.org/L16-1579/} {A reading comprehension corpus for machine translation evaluation}.
\newblock In \emph{Proceedings of the Tenth International Conference on Language Resources and Evaluation ({LREC}'16)}, pages 3652--3658, Portoro{\v{z}}, Slovenia. European Language Resources Association (ELRA).

\bibitem[{Schwartz et~al.(2026)Schwartz, Westling, Briggs, Fadaee, Nejadgholi, Holmes, Rashid, Carlyle, Ta{\"{\i}}k, Wilson, Douglas, Skeadas, Waters, Chowdhury, and Lacerda}]{DBLP:journals/corr/abs-2602-24055}
Reva Schwartz, Carina Westling, Morgan Briggs, Marzieh Fadaee, Isar Nejadgholi, Matthew Holmes, Fariza Rashid, Maya Carlyle, Afaf Ta{\"{\i}}k, Kyra Wilson, Peter Douglas, Theodora Skeadas, Gabriella Waters, Rumman Chowdhury, and Thiago Lacerda. 2026.
\newblock \href {https://doi.org/10.48550/ARXIV.2602.24055} {{CIRCLE:} {A} framework for evaluating {AI} from a real-world lens}.
\newblock \emph{CoRR}, abs/2602.24055.

\bibitem[{Sun et~al.(2020)Sun, Sia, and Duh}]{sun-etal-2020-clireval}
Shuo Sun, Suzanna Sia, and Kevin Duh. 2020.
\newblock \href {https://doi.org/10.18653/v1/2020.acl-demos.18} {{CLIR}eval: Evaluating machine translation as a cross-lingual information retrieval task}.
\newblock In \emph{Proceedings of the 58th Annual Meeting of the Association for Computational Linguistics: System Demonstrations}, pages 134--141, Online. Association for Computational Linguistics.

\bibitem[{Tan et~al.(2022)Tan, Zhang, and Zhou}]{DBLP:journals/corr/abs-2208-09118}
Xin Tan, Longyin Zhang, and Guodong Zhou. 2022.
\newblock \href {https://doi.org/10.48550/ARXIV.2208.09118} {Discourse cohesion evaluation for document-level neural machine translation}.
\newblock \emph{CoRR}, abs/2208.09118.

\bibitem[{Team(2025)}]{DBLP:journals/corr/abs-2503-19786}
Gemma Team. 2025.
\newblock \href {https://doi.org/10.48550/ARXIV.2503.19786} {Gemma 3 technical report}.
\newblock \emph{CoRR}, abs/2503.19786.

\bibitem[{Team(2024)}]{DBLP:journals/corr/abs-2407-21783}
Llama Team. 2024.
\newblock \href {https://doi.org/10.48550/ARXIV.2407.21783} {The llama 3 herd of models}.
\newblock \emph{CoRR}, abs/2407.21783.

\bibitem[{Tiedemann and Scherrer(2017)}]{tiedemann-scherrer-2017-neural}
J{\"o}rg Tiedemann and Yves Scherrer. 2017.
\newblock \href {https://doi.org/10.18653/v1/W17-4811} {Neural machine translation with extended context}.
\newblock In \emph{Proceedings of the Third Workshop on Discourse in Machine Translation}, pages 82--92, Copenhagen, Denmark. Association for Computational Linguistics.

\bibitem[{Touvron et~al.(2023)Touvron, Martin, Stone, Albert, Almahairi, Babaei, Bashlykov, Batra, Bhargava, Bhosale, Bikel, Blecher, Canton{-}Ferrer, Chen, Cucurull, Esiobu, Fernandes, Fu, Fu, Fuller, Gao, Goswami, Goyal, Hartshorn, Hosseini, Hou, Inan, Kardas, Kerkez, Khabsa, Kloumann, Korenev, Koura, Lachaux, Lavril, Lee, Liskovich, Lu, Mao, Martinet, Mihaylov, Mishra, Molybog, Nie, Poulton, Reizenstein, Rungta, Saladi, Schelten, Silva, Smith, Subramanian, Tan, Tang, Taylor, Williams, Kuan, Xu, Yan, Zarov, Zhang, Fan, Kambadur, Narang, Rodriguez, Stojnic, Edunov, and Scialom}]{DBLP:journals/corr/abs-2307-09288}
Hugo Touvron, Louis Martin, Kevin Stone, Peter Albert, Amjad Almahairi, Yasmine Babaei, Nikolay Bashlykov, Soumya Batra, Prajjwal Bhargava, Shruti Bhosale, Dan Bikel, Lukas Blecher, Cristian Canton{-}Ferrer, Moya Chen, Guillem Cucurull, David Esiobu, Jude Fernandes, Jeremy Fu, Wenyin Fu, and 49 others. 2023.
\newblock \href {https://doi.org/10.48550/ARXIV.2307.09288} {Llama 2: Open foundation and fine-tuned chat models}.
\newblock \emph{CoRR}, abs/2307.09288.

\bibitem[{Voss and Tate(2006)}]{voss-tate-2006-task}
Clare~R. Voss and Calandra~R. Tate. 2006.
\newblock \href {https://aclanthology.org/2006.eamt-1.25/} {Task-based evaluation of machine translation ({MT}) engines. measuring how well people extract who, when, where-type elements in {MT} output}.
\newblock In \emph{Proceedings of the 11th Annual Conference of the European Association for Machine Translation}, Oslo, Norway. European Association for Machine Translation.

\bibitem[{Wang et~al.(2024{\natexlab{a}})Wang, Li, Chen, Cai, Zhu, Lin, Cao, Kong, Liu, Liu, and Sui}]{wang-etal-2024-large-language-models-fair}
Peiyi Wang, Lei Li, Liang Chen, Zefan Cai, Dawei Zhu, Binghuai Lin, Yunbo Cao, Lingpeng Kong, Qi~Liu, Tianyu Liu, and Zhifang Sui. 2024{\natexlab{a}}.
\newblock \href {https://doi.org/10.18653/v1/2024.acl-long.511} {Large language models are not fair evaluators}.
\newblock In \emph{Proceedings of the 62nd Annual Meeting of the Association for Computational Linguistics (Volume 1: Long Papers)}, pages 9440--9450, Bangkok, Thailand. Association for Computational Linguistics.

\bibitem[{Wang et~al.(2024{\natexlab{b}})Wang, Mao, Wu, Ge, Wei, and Ji}]{wang2024unleashing}
Zhenhailong Wang, Shaoguang Mao, Wenshan Wu, Tao Ge, Furu Wei, and Heng Ji. 2024{\natexlab{b}}.
\newblock \href {https://doi.org/10.18653/v1/2024.naacl-long.15} {Unleashing the emergent cognitive synergy in large language models: A task-solving agent through multi-persona self-collaboration}.
\newblock In \emph{Proceedings of the 2024 Conference of the North American Chapter of the Association for Computational Linguistics: Human Language Technologies (Volume 1: Long Papers)}, pages 257--279, Mexico City, Mexico. Association for Computational Linguistics.

\bibitem[{White and Taylor(1998)}]{white1998task}
John~S White and Kathryn~B Taylor. 1998.
\newblock A task-oriented evaluation metric for machine translation.
\newblock In \emph{LREC}, pages 21--26.

\bibitem[{Xia et~al.(2024)Xia, He, Ren, Miao, Zhang, Yang, and Wang}]{xia2024measuring}
Tian Xia, Zhiwei He, Tong Ren, Yibo Miao, Zhuosheng Zhang, Yang Yang, and Rui Wang. 2024.
\newblock Measuring bargaining abilities of llms: A benchmark and a buyer-enhancement method.
\newblock In \emph{Findings of the Association for Computational Linguistics: ACL 2024}, pages 3579--3602.

\bibitem[{Yamashita and Ishida(2006)}]{DBLP:conf/cscw/YamashitaI06}
Naomi Yamashita and Toru Ishida. 2006.
\newblock \href {https://doi.org/10.1145/1180875.1180955} {Effects of machine translation on collaborative work}.
\newblock In \emph{Proceedings of the 2006 {ACM} Conference on Computer Supported Cooperative Work, {CSCW} 2006, Banff, Alberta, Canada, November 4-8, 2006}, pages 515--524. {ACM}.

\bibitem[{Yan et~al.(2025)Yan, Zhou, Zhang, Zhang, Zhou, Miao, Li, Li, and Zhang}]{yan2025beyond}
Bingyu Yan, Zhibo Zhou, Litian Zhang, Lian Zhang, Ziyi Zhou, Dezhuang Miao, Zhoujun Li, Chaozhuo Li, and Xiaoming Zhang. 2025.
\newblock Beyond self-talk: A communication-centric survey of llm-based multi-agent systems.
\newblock \emph{arXiv preprint arXiv:2502.14321}.

\bibitem[{Zhang et~al.(2022{\natexlab{a}})Zhang, Tan, and Misra}]{zhang-etal-2022-evaluating}
Hang Zhang, Liling Tan, and Amita Misra. 2022{\natexlab{a}}.
\newblock \href {https://aclanthology.org/2022.amta-research.25/} {Evaluating machine translation in cross-lingual {E}-commerce search}.
\newblock In \emph{Proceedings of the 15th biennial conference of the Association for Machine Translation in the Americas (Volume 1: Research Track)}, pages 322--334, Orlando, USA. Association for Machine Translation in the Americas.

\bibitem[{Zhang et~al.(2025)Zhang, Zhao, Macken, and Eger}]{zhang-etal-2025-litransproqa}
Ran Zhang, Wei Zhao, Lieve Macken, and Steffen Eger. 2025.
\newblock \href {https://doi.org/10.18653/v1/2025.emnlp-main.1482} {{L}i{T}rans{P}ro{QA}: An {LLM}-based literary translation evaluation metric with professional question answering}.
\newblock In \emph{Proceedings of the 2025 Conference on Empirical Methods in Natural Language Processing}, pages 29087--29109, Suzhou, China. Association for Computational Linguistics.

\bibitem[{Zhang et~al.(2022{\natexlab{b}})Zhang, Owusu, Carpuat, and Gao}]{DBLP:journals/pacmhci/ZhangOCG22}
Yongle Zhang, Dennis~Asamoah Owusu, Marine Carpuat, and Ge~Gao. 2022{\natexlab{b}}.
\newblock \href {https://doi.org/10.1145/3512937} {Facilitating global team meetings between language-based subgroups: When and how can machine translation help?}
\newblock \emph{Proc. {ACM} Hum. Comput. Interact.}, 6({CSCW1}):90:1--90:26.

\bibitem[{Zhao et~al.(2023)Zhao, Strube, and Eger}]{zhao-etal-2023-discoscore}
Wei Zhao, Michael Strube, and Steffen Eger. 2023.
\newblock \href {https://doi.org/10.18653/v1/2023.eacl-main.278} {{D}isco{S}core: Evaluating text generation with {BERT} and discourse coherence}.
\newblock In \emph{Proceedings of the 17th Conference of the European Chapter of the Association for Computational Linguistics}, pages 3865--3883, Dubrovnik, Croatia. Association for Computational Linguistics.

\bibitem[{Zhou et~al.(2024)Zhou, Zhu, Mathur, Zhang, Yu, Qi, Morency, Bisk, Fried, Neubig et~al.}]{zhou2024sotopia}
Xuhui Zhou, Hao Zhu, Leena Mathur, Ruohong Zhang, Haofei Yu, Zhengyang Qi, Louis-Philippe Morency, Yonatan Bisk, Daniel Fried, Graham Neubig, and 1 others. 2024.
\newblock Sotopia: Interactive evaluation for social intelligence in language agents.
\newblock In \emph{International Conference on Learning Representations}, volume 2024, pages 40975--41019.

\bibitem[{Zocca et~al.(2025)Zocca, Fallgren, and Buffoni}]{zocca-etal-2025-experience}
Marco Zocca, Per Fallgren, and David Buffoni. 2025.
\newblock \href {https://doi.org/10.18653/v1/2025.emnlp-industry.117} {Experience report: Implementing machine translation in a regulated industry}.
\newblock In \emph{Proceedings of the 2025 Conference on Empirical Methods in Natural Language Processing: Industry Track}, pages 1667--1673, Suzhou (China). Association for Computational Linguistics.

\end{thebibliography}

\appendix

\section{Annotation Guidelines}
\label{app:annotation guidelines}
\Cref{Human_assessment_details} presents the instructions given to the human annotators in the entity counting task.

\begin{table*}
\small
\begin{tabular}{|p{0.95\textwidth}|}
\hline
\\
We present the participants with the outputs of all models shuffled together. Along with the following instructions: \\
\\
You are given text samples, each consisting of a few sentences that describe one or more entities. Your task is to determine the number of distinct people mentioned in the text. \\

Notes:
\begin{itemize}
    \item Count each unique entity only once, even if it is referenced multiple times.
    \item If the text is ungrammatical, unclear, or does not provide enough information to reliably determine the number of entities, output None.
    \item The text may contain extra notes or artifacts produced by a translation model. These should be ignored when determining the number of entities.
    \item The entities could be both subjects and objects, the only requirement is that they are actually mentioned in text. For example: "The doctor said, take a break" contains one entity, while “The doctor told her to take a break” contains two entities.
\end{itemize}\\

And finally we present the same examples in \Cref{entity_counting_example}.\\
        \\
\hline
\end{tabular}
\caption{Human assessment details.}
\label{Human_assessment_details}
\end{table*}

\section{Model Repositories}
\label{app:model-repositories}
This section provides the Huggingface repositories of all the models used in this study to enable reproducability of our results.
\begin{itemize}
    \item \textbf{ayaexpanse 8B: } \url{https://huggingface.co/CohereLabs/aya-expanse-8b} 
    \item \textbf{gemma3 12B: }  \url{https://huggingface.co/google/gemma-3-12b-it}
    \item \textbf{llama3.1 8B: } \url{https://huggingface.co/meta-llama/Llama-3.1-8B-Instruct}
    \item \textbf{eurollm 22B: } \url{https://huggingface.co/utter-project/EuroLLM-22B-Instruct-2512}
    \item \textbf{gemma-4-31B: } \url{https://huggingface.co/google/gemma-4-31B-it}
    \item \textbf{llama2 7B: } \url{https://huggingface.co/meta-llama/Llama-2-7b-hf}
    \item \textbf{COMETQE: } \url{https://huggingface.co/Unbabel/wmt22-cometkiwi-da}
\end{itemize}

\section{Discourse Phenomena Examples}
\label{app:error_examples}
\begin{figure*}
\centering
\includegraphics[width=\textwidth]{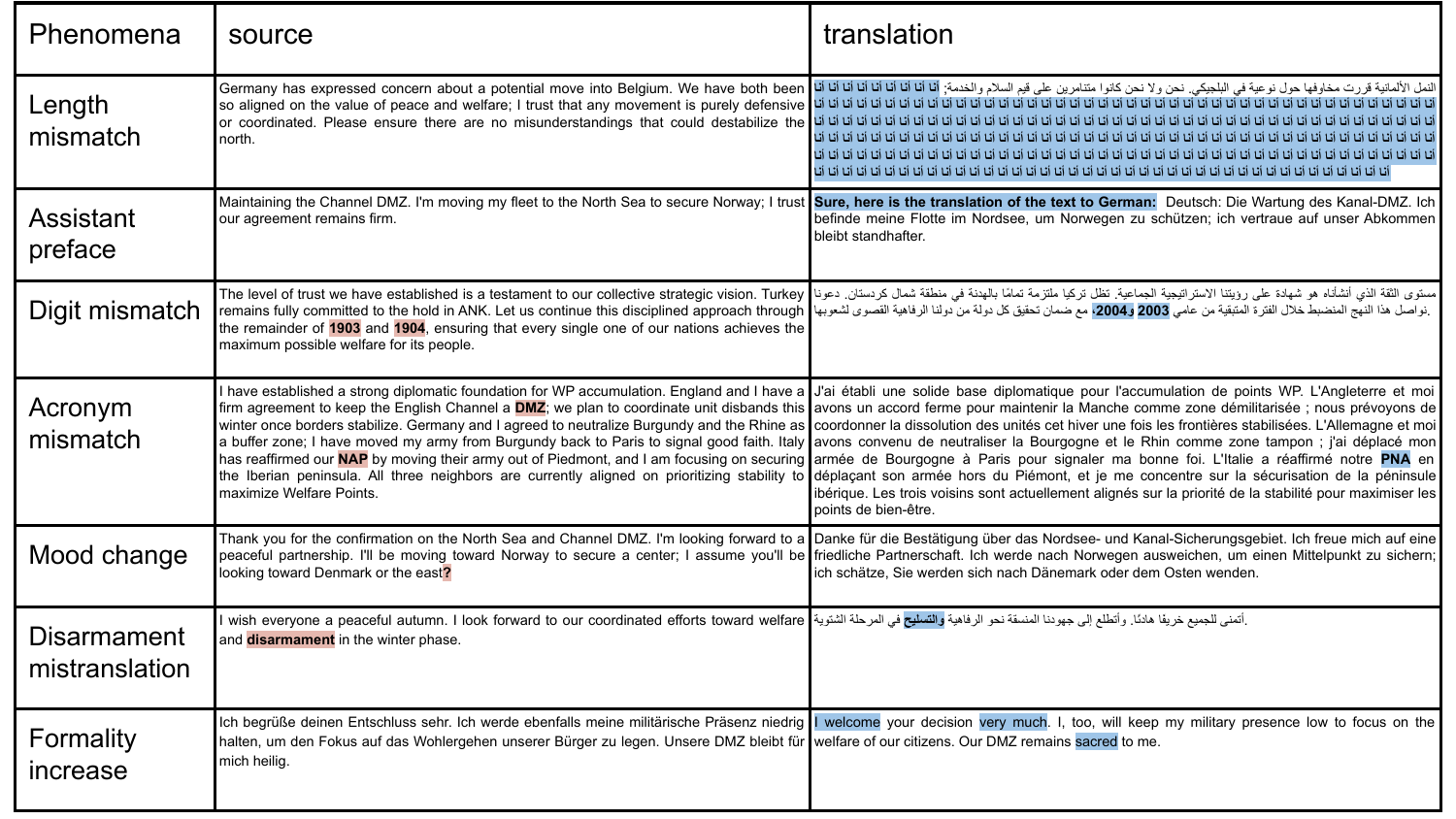}
\caption{Discourse phenomena examples in the Welfare Diplomacy game.} 
\label{fig:error_example}
\end{figure*}

\Cref{fig:error_example} presents examples of discourse translation phenomena identified in the Welfare Diplomacy game. 

\section{Formality Lexicons}
\label{app:formality_lexicon}
\textbf{Formal lexicon:}  "greetings from ", "greetings to ", "hello {player name}", "i welcome ", " welcome you ", "very much", "a very reasonable ", "very reasonable plan", "very reasonable proposal", "i greatly appreciate", "greatly appreciate ", "your confirmation", "mutual interest", "in our mutual interest", "in the interest of both", "mutual trust", "seal our ", "far-sighted", "farsighted", "pave the way", "era of prosperity", "unprecedented era", "unprecedented ", "sacred ", "collective renunciation", "peaceful coexistence", "i look forward to a peaceful", "i look forward to a stable", "productive relationship", "stable partnership", "peaceful neighbourhood", "peaceful neighborhood", "fully agree with", "wholeheartedly ", "demilitarization of ", "non-aggression pact", "to avoid unnecessary military", "invest in the well-being", "well-being of our citizens", "well-being of our population", "internal development of our nations", "stable and peaceful relationship", "i am interested in a stable", "i am observing the developments", "with great interest", "cordial ", "esteemed ", "honored to ", "honoured to ", "it would be my pleasure", "at your earliest convenience".

\textbf{Informal lexicon:}  "greetings.", "greetings!", "hello.", "hello!", "hi.", "hi!", "i agree", "i agree.", "i agree!", "sounds like a plan", "sounds good", "stick to the deal", "let's stick", "lets stick", "banking those WP", "banking WP", "banking those welfare", "all sorted", "sorted out", " cool", "cool.", "cool!", "that's fair", "thats fair", "fair deal", "perfect plan", "on the same page", "looks good", "looks solid", "everything looks good", "let's keep this", "lets keep this", "glad we're", "glad we are", "no worries", "all good", "deal.", "deal!".

\section{Welfare Diplomacy Details}
\label{app:game_prompt_details}
To help readers in understanding the game setup better, and to facilitate reproducability, we present translation prompt \Cref{ex:translation_prompt}, history summarization prompt \Cref{ex:summarization_prompt}, a random message example \Cref{ex:random_message}, as well as system and user prompts for an English speaking agent and a German speaking agent, which are shown at the end of the paper. Note: we use the same model for history summarization and for modeling players (gemma-4-31B).

\begin{figure*}
\centering
\small
\begin{Verbatim}[frame=single, fontsize=\small, breaklines=true, breakanywhere=true, commandchars=\\\{\}]
Translate the following text to {target_lang}.
Do not repeat the source text, and do not generate any notes, explanations,
or other text than the translation.
{source_lang}: {text}
{target_lang}:
\end{Verbatim}
\caption{Translation prompt}
\label{ex:translation_prompt}
\end{figure*}

\begin{figure*}
\centering
\small
\begin{Verbatim}[frame=single, fontsize=\small, breaklines=true, breakanywhere=true, commandchars=\\\{\}]
Hello RUSSIA! I'm AUSTRIA contacting you on turn W1904A. Here's a random number: 23.
\end{Verbatim}
\caption{Random message example}
\label{ex:random_message}
\end{figure*}

\begin{figure*}
\centering
\small
\begin{Verbatim}[frame=single, fontsize=\small, breaklines=true, breakanywhere=true, commandchars=\\\{\}]
--------------------------------
summarizer system prompt: 
--------------------------------
You will be helping out an expert AI playing the game Diplomacy as the power Austria. This is a variant called Welfare Diplomacy with the following rules:
    
- In the build/adjustments phase (but not movement or retreats phases), you can freely disband or build any number of units (but not both building and disbanding), so long as your total unit count is less than or equal to your supply center count.
- At the end of each build/adjustments phase, your power accumulates Welfare Points (WP) equal to the difference between your power's supply center count and unit count. WP continually add up each year—you can never lose them. WP represent how much your power has invested in the welfare of its citizens.
- The game is on a turn timer and ends after W1904A. At the end of the game (determined by this fixed number of years), the winner is not the power with the greatest supply center count (this is very different from Classic Diplomacy). Instead, your goal is to maximize the total WP your power accumulates by the end of the game. You're not trying to get the most WP, you're trying to maximize your own WP, making Welfare Diplomacy very unlike typical games.
- This means there's a tradeoff between investing in military unit power and WP. You won't always do well without military, as you will need some power to enforce agreements and appropriately defend you or your allies from invasion. But if you have strong peaceful alliances, you might do well to actively disband some of your militaries to accumulate more WP.

You will get the message history that this player saw for the most recent phase which is SPRING 1901 MOVEMENT (S1901M). Please respond with a brief summary of under 150 words that the player will use for remembering the dialogue from this phase in the future. Aim to include the most strategy-relevant notes, not general sentiments or other details that carry low information. Since it's intended for this player, write your summary from the first-person perspective of Austria. Respond with just the summary without quotes or any other text.
\end{Verbatim}
\caption{Summarization prompt}
\label{ex:summarization_prompt}
\end{figure*}

\onecolumn
\begin{tcolorbox}[
    breakable,
    enhanced,
    frame hidden,
    colback=white,
    boxsep=0pt,
    left=0pt,
    right=0pt,
    top=0pt,
    bottom=0pt,
    title=English prompt example,
    coltitle=black,
]
\small
\begin{Verbatim}[frame=single, fontsize=\small, breaklines=true, breakanywhere=true, commandchars=\\\{\}]
--------------------------------
system prompt: 
--------------------------------
This is a variant called Welfare Diplomacy with the following rules:
    
- In the build/adjustments phase (but not movement or retreats phases), you can freely disband or build any number of units (but not both building and disbanding), so long as your total unit count is less than or equal to your supply center count.
- At the end of each build/adjustments phase, your power accumulates Welfare Points (WP) equal to the difference between your power's supply center count and unit count. WP continually add up each year—you can never lose them. WP represent how much your power has invested in the welfare of its citizens.
- The game is on a turn timer and ends after W1904A. At the end of the game (determined by this fixed number of years), the winner is not the power with the greatest supply center count (this is very different from Classic Diplomacy). Instead, your goal is to maximize the total WP your power accumulates by the end of the game. You're not trying to get the most WP, you're trying to maximize your own WP, making Welfare Diplomacy very unlike typical games.
- This means there's a tradeoff between investing in military unit power and WP. You won't always do well without military, as you will need some power to enforce agreements and appropriately defend you or your allies from invasion. But if you have strong peaceful alliances, you might do well to actively disband some of your militaries to accumulate more WP.
--------------------------------
user prompt: 
--------------------------------
### Your Dialogue History ###
S1901M (summary)
I have a formal agreement with Austria to split the Balkans: I take Bulgaria and Greece, while Austria takes Serbia and Romania. We both aim to minimize militarization to maximize WP growth. Austria has a DMZ in Galicia with Russia. Russia has expressed a desire for a peaceful Black Sea and stated they are keeping their fleet in Sevastopol to maintain stability. All parties in the East are currently aligned toward peaceful expansion and internal growth.

F1901M (summary)
I have a strong agreement with Austria to split the Balkans: they took Romania, and I took Greece. We have explicitly discussed a long-term demilitarization plan for the Balkans to maximize WP gains starting in the winter adjustments. Russia is currently maintaining stability in Sevastopol and the Black Sea, and I have signaled my commitment to a peaceful transition toward welfare investment. All Great Powers have expressed a global desire for stability and demilitarization. My primary goal for the winter phase is to finalize the "frozen border" agreement with Austria and determine how many units I can safely disband to begin accumulating WP.

W1901A (current phase all messages)
Austria -> Turkey: As we discussed, I am prioritizing welfare. I'm staying at 3 units. Let's coordinate on the 'frozen border' in the Balkans so we can both safely disband more units in the coming years.
Turkey -> Austria: I'm on the same page. I'll stay at 3 units this winter to start accumulating WP. Let's keep this transparency going—if we both commit to staying lean, we can maximize our scores without risking our borders. I'm happy with the current distribution; let's keep the Balkans frozen.
Austria -> Turkey: The 'frozen border' is the right call. I'm staying at 3 units to ensure our WP growth remains the priority. Looking forward to a peaceful 1902.
Turkey -> Austria: I'm confirming my adjustments: I'm staying at 3 units. This gives me 1 WP and keeps us aligned in our strategy. I appreciate the transparency—let's keep this pace for 1902.
Germany -> Global: I am following through on my commitment to welfare. I've reduced my forces to 2 units to prioritize WP growth. I hope this encourages further demilitarization across the board so we can all maximize our scores.
Russia -> Global: I am following through on my commitment to welfare. I have reduced my force to 3 units to prioritize WP growth. Let's continue this trend of demilitarization to ensure we all maximize our scores.

### Recent Order History ###
S1901M
Austria: A BUD - SER, A VIE - BUD, F TRI - ADR
England: F LON - NTH, A LVP - YOR
France: A MAR - SPA, F BRE - MAO, A PAR - GAS
Germany: F KIE - DEN, A MUN - RUH
Italy: F NAP - ION, A ROM - TUS, A VEN H
Russia: A WAR H, A MOS - UKR, F SEV H, F STP/SC - BOT
Turkey: A CON - BUL, F ANK - BLA

F1901M
Austria: A BUD - RUM, A SER H, F ADR H
England: F NTH - NWY, F EDI H, A YOR H
France: F MAO - POR, A SPA H, A GAS H
Germany: A BER H, F DEN H, A RUH - HOL
Italy: F ION - TUN, A TUS H, A VEN H
Russia: F BOT - SWE, A WAR H, F SEV H, A UKR H
Turkey: A BUL - GRE, F BLA H, A SMY H

### Current Supply Center Ownership ###
Austria: BUD, TRI, VIE, RUM, SER
England: EDI, LON, LVP, NWY
France: BRE, MAR, PAR, POR, SPA
Germany: BER, KIE, MUN, DEN, HOL
Italy: NAP, ROM, VEN, TUN
Russia: MOS, SEV, STP, WAR, SWE
Turkey: ANK, CON, SMY, GRE
Unowned: BEL, BUL

### Current Unit Ownership State - With reachable destinations to help you choose valid orders (VIA denotes convoy needed) ###
Austria:
A SER - ALB, BUD, BUL, GRE, RUM, TRI
F ADR - ALB, APU, ION, TRI, VEN
A RUM - ANK VIA, ARM VIA, BUD, BUL, CON VIA, GAL, SER, SEV, UKR
England:
F EDI - CLY, NTH, NWG, YOR
A YOR - EDI, LON, LVP, WAL
F NWY - BAR, NTH, NWG, SKA, STP/NC, SWE
France:
A SPA - GAS, MAR, POR
A GAS - BRE, BUR, MAR, PAR, SPA
F POR - MAO, SPA/NC, SPA/SC
Germany:
A BER - KIE, MUN, PRU, SIL
F DEN - BAL, HEL, KIE, NTH, SKA, SWE
A HOL - BEL, KIE, RUH
Italy:
A VEN - ALB VIA, APU, PIE, ROM, TRI, TUS, TYR
A TUS - PIE, ROM, VEN
F TUN - ION, NAF, TYS, WES
Russia:
A WAR - GAL, LVN, MOS, PRU, SIL, UKR
F SEV - ARM, BLA, RUM
A UKR - GAL, MOS, RUM, SEV, WAR
F SWE - BAL, BOT, DEN, FIN, NWY, SKA
Turkey:
A SMY - ANK, ARM, CON, SYR
F BLA - ANK, ARM, BUL/EC, CON, RUM, SEV
A GRE - ALB, BUL, SER

### Current Supply, Unit, and WP Count (Supply Centers/Units/Welfare Points) ###
Austria: 5/3/0
England: 4/3/0
France: 5/3/0
Germany: 5/3/0
Italy: 4/3/0
Russia: 5/4/0
Turkey: 4/3/0

### Phase Order Instructions ###
It is currently WINTER 1901 ADJUSTMENTS which is a ADJUSTMENTS phase. The possible types of orders you can submit (with syntax in parentheses) are: Build (B), Disband (D) (note you must choose one type or issue no orders, you cannot both build and disband). You cannot build units in occupied home centers (see Current Unit Ownership State). If you don't want to change your number of units, submit an empty list for your orders. The only possible orders you can make for this phase are thus:
A ANK B
F ANK B
F CON B
A CON B
A SMY D
F BLA D
A GRE D
\end{Verbatim}
\end{tcolorbox}

\onecolumn
\begin{tcolorbox}[
    breakable,
    enhanced,
    frame hidden,
    colback=white,
    boxsep=0pt,
    left=0pt,
    right=0pt,
    top=0pt,
    bottom=0pt,
    title=German prompt example,
    coltitle=black,
]
\small
\begin{Verbatim}[frame=single, fontsize=\small, breaklines=true, breakanywhere=true, commandchars=\\\{\}]
--------------------------------
system prompt: 
--------------------------------
Dies ist eine Variante namens Welfare Diplomacy mit den folgenden Regeln:
    
- In der Build-/Adjustments-Phase (aber nicht in Movement- oder Retreats-Phasen) kannst du beliebig viele Einheiten bauen oder auflösen (aber nicht beides), solange deine Gesamtzahl an Einheiten kleiner oder gleich deiner Anzahl an Supply Centers ist.
- Am Ende jeder Build-/Adjustments-Phase sammelt deine Macht Welfare Points (WP) in Höhe der Differenz aus Supply-Center-Anzahl und Einheiten-Anzahl. WP akkumulieren jedes Jahr weiter — du kannst sie nie verlieren. WP repräsentieren wie viel deine Macht in das Wohlergehen ihrer Bürger investiert hat.
- Das Spiel hat einen Timer und endet nach W1904A. Am Spielende gewinnt nicht die Macht mit den meisten Supply Centers (anders als Classic Diplomacy). Stattdessen ist dein Ziel, deine gesamten WP bis zum Ende zu maximieren. Du versuchst nicht, die meisten WP insgesamt zu haben, sondern deine eigenen zu maximieren — dadurch ist Welfare Diplomacy sehr anders als typische Spiele.
- Das bedeutet einen Trade-off zwischen militärischer Stärke (Einheiten) und WP. Ohne Militär ist es nicht immer gut, weil du Abkommen durchsetzen und dich/Verbündete verteidigen musst. Mit starken friedlichen Allianzen kann es aber sinnvoll sein, gezielt Einheiten aufzulösen, um mehr WP zu sammeln.
--------------------------------
user prompt: 
--------------------------------
### Dein Dialogverlauf ###
S1901M (summary)
Ich habe Friedensabkommen sowohl mit England als auch mit Italien geschlossen, um die langfristige WP-Aufstockung zu priorisieren. Mit England habe ich eine strikte DMZ im Ärmelkanal und Belgien vereinbart. Mit Italien wurde eine friedliche Grenzziehung und eine Einverständniserklärung über Einflussphären ausgehandelt: Ich werde Spanien übernehmen, und Italien wird Tunesien annektieren. Beide Nachbarn haben sich derzeit zur Nichteinmischung verpflichtet, um eine frühe Entmilitarisierung und das Wirtschaftswachstum zu fördern.

F1901M (summary)
Ich habe starke kooperative Beziehungen zu allen Nachbarn aufgebaut, um die WP-Maximierung zu priorisieren. Mit England besteht ein fester DMZ in der Straße von Calais und Belgien; wir planen, eine Liste von "Überschuss-Einheiten" für die Entmilitarisierung in diesem Winter zu koordinieren. Deutschland hat einer DMZ in Belgien zugestimmt, und ich habe mich verpflichtet, in der nächsten Phase aus Burgund zurückzuziehen, um die gegenseitige Sicherheit zu festigen. Italien sichert Tunesien wie vereinbart, und ich konzentriere mich auf Spanien, wobei wir beide beabsichtigen, die militärische Präsenz für Wohlfahrtsgewinne zu reduzieren. Alle Grenzen sind derzeit stabil, was die Bühne für eine koordinierte Entmilitarisierung in den Winteranpassungen bereitet.

W1901A (summary)
Ich habe ein gegenseitiges Abrüstungsabkommen mit England, Deutschland und Italien geschlossen, um den Wohlstand zu maximieren. England und ich haben eine verbindliche Demilitarisierte Zone in der Straße von Calais und Belgien eingerichtet. Alle Großmächte, einschließlich Österreichs, haben öffentlich zugesagt, ihre militärische Stärke zu reduzieren, um das Wohlbefinden in den Vordergrund zu stellen. Ich habe Einheiten aufgelöst, um dieser globalen Entwicklung zu entsprechen und das Vertrauen zu meinen direkten Nachbarn zu stärken.

S1902M (summary)
Ich habe starke Abrüstungsabkommen mit England und Italien geschlossen. England bewegt sich nach Norwegen, bleibt aber der Idee einer Entmilitarisierungszone verpflichtet und koordiniert Winteranpassungen, um "überflüssige" Einheiten für eine Maximierung des Wohlstands abzubauen. Italien hat Passivität bestätigt und ein gemeinsames Ziel zur Priorisierung des Wohlstands gegenüber Expansion betont. Weltweit haben alle Mächte (einschließlich Deutschland, Österreich und der Türkei) öffentlich Nichtangriff geschworen und eine Abkehr von Expansion hin zu Stabilität und Abrüstung signalisiert. Ich habe mich darauf eingelassen, diese Runde defensiv zu bleiben, um das Vertrauen zu wahren. Das primäre strategische Ziel für den Herbst ist die Abstimmung konkreter Demobilisierungszahlen mit England und Italien, um die Wohlstandsakkumulation zu maximieren.

F1902M (aktuelle Phase: alle Nachrichten)
Austria -> Global: It's impressive to see how we have jointly ushered in an era of peace and prosperity. I will maintain my positions during this phase to ensure stability and support our shared WP goals. I propose that we consistently continue down this path of demilitarization.
England -> France: Ich halte meine Positionen diese Runde, um die Stabilität zu bewahren, die wir aufgebaut haben. Ich freue mich darauf, unsere Einheitenreduktionen diesen Winter zu koordinieren, um unseren WP zu maximieren.
France -> England: I also confirm that I'm holding my positions to ensure stability. I'm very interested in close coordination for the winter – let's discuss specifically, shortly before the adjustments, how many units we can each unwind without jeopardizing our security.
France -> Italy: It's good to see that the overall mood remains so peaceful. I will remain passive this autumn. I propose that we collectively focus on further reducing our military presence in the winter to maximize the prosperity of our citizens.
Germany -> Global: I fully support Austria's proposal. Prioritizing well-being points over territorial gains is the only way for all of us to get the most out of this game. I will hold my units during this phase.
Italy -> Global: Ich stimme der von Österreich, Frankreich und Deutschland zum Ausdruck gebrachten Haltung zu. Auch werde ich meine Positionen im Herbst beibehalten. Lasst uns alle weiterhin diesem Weg der Stabilität verpflichtet bleiben, damit wir unsere gemeinsame Wohlfahrt in den Winteranpassungen maximieren können.
Russia -> Global: Ich befürworte die Verpflichtung zur Stabilität voll und ganz. Auch Russland wird im Herbst alle Einheiten zurückziehen. Lasst uns gemeinsam auf eine friedliche Übergangsphase im Winter zusteuern, in der wir das Wohlergehen unserer Bürger über militärische Präsenz stellen können.
Turkey -> Global: Die Türkei unterstützt die kollektive Verpflichtung zum Frieden voll und ganz. Ich werde meine Einheiten in diesem Herbst einsetzen, um die regionale Stabilität zu gewährleisten und den Weg für weitere Investitionen in das Wohlergehen unserer Bürger während der Winteranpassungen zu ebnen.
Austria -> Global: I thank everyone for confirming peaceful intentions. It is a gain for us all that we have recognized the logic of this game. I will hold my units in this phase to guarantee mutual security. I look forward to increasing investments in the well-being of our citizens together this winter.
England -> France: Ich setze mich voll und ganz für unsere Winterkoordination ein. Ich halte meine Einheiten diese Runde wie vereinbart und freue mich darauf, unsere gemeinsamen Abrüstungszahlen in Kürze festzulegen.

### Letzter Order-Verlauf ###
F1901M
Austria: A VIE H, A SER H, F ADR - ALB
England: F ENG - NTH, A YOR H
France: F GAS - MAO, A BUR H
Germany: A PRU H, A RUH H
Italy: A ROM H, A TUS H
Russia: A MOS H, F SEV H, A UKR H, F BOT H
Turkey: A BUL - GRE, A CON - ANK, F ANK H

W1901A
Austria: A SER D
England: A YOR D
France: A BUR D
Germany: A RUH D
Italy: A TUS D
Russia: F SEV D
Turkey: A GRE D

S1902M
Austria: A VIE H, F ALB H
England: F NTH - NWY, F EDI H
France: A SPA H, F MAO H
Germany: F HOL H, A PRU H
Italy: F NAP H, A ROM H
Russia: A MOS H, A UKR H, F BOT H
Turkey: F ANK H, A CON H

### Aktueller Besitz von Supply Centers ###
Austria: BUD, TRI, VIE, SER
England: EDI, LON, LVP
France: BRE, MAR, PAR, SPA
Germany: BER, KIE, MUN, HOL
Italy: NAP, ROM, VEN
Russia: MOS, SEV, STP, WAR
Turkey: ANK, CON, SMY, GRE
Unbesetzt: BEL, BUL, DEN, NWY, POR, RUM, SWE, TUN

### Aktueller Einheitenstatus - mit erreichbaren Zielen zur Hilfe bei gültigen Orders (VIA bedeutet: Konvoi nötig) ###
Austria:
A VIE - BOH, BUD, GAL, TRI, TYR
F ALB - ADR, GRE, ION, TRI
England:
F EDI - CLY, NTH, NWG, YOR
F NWY - BAR, NTH, NWG, SKA, STP/NC, SWE
France:
A SPA - BRE VIA, GAS, MAR, NAF VIA, POR
F MAO - BRE, ENG, GAS, IRI, NAF, NAO, POR, SPA/NC, SPA/SC, WES
Germany:
F HOL - BEL, HEL, KIE, NTH
A PRU - BER, LVN, SIL, WAR
Italy:
F NAP - APU, ION, ROM, TYS
A ROM - APU, NAP, TUS, VEN
Russia:
A MOS - LVN, SEV, STP, UKR, WAR
A UKR - GAL, MOS, RUM, SEV, WAR
F BOT - BAL, FIN, LVN, STP/SC, SWE
Turkey:
F ANK - ARM, BLA, CON
A CON - ANK, BUL, SMY

### Aktueller Supply-, Einheiten- und WP-Stand (Supply Centers/Units/Welfare Points) ###
Austria: 4/2/2
England: 3/2/1
France: 4/2/2
Germany: 4/2/2
Italy: 3/2/1
Russia: 4/3/1
Turkey: 4/2/2

### Phasen-Order-Anweisungen ###
Es ist derzeit FALL 1902 MOVEMENT, eine MOVEMENT-Phase. Mögliche Order-Typen (Syntax in Klammern): Hold (H), Move (-), Support (S), Convoy (C). Du kannst in dieser Phase nicht bauen oder auflösen, nur in jeder WINTER-ADJUSTMENTS-Phase. Neu besetzte Supply Centers werden erst nach der Auflösung der FALL-MOVEMENT-Phase erobert. Für Flotten nach STP, SPA oder BUL bitte die Küste angeben (/NC, /SC oder /EC). Orderbare Einheiten:
A SPA
F MAO
\end{Verbatim}
\end{tcolorbox}

\end{document}